\documentclass{bmvc2k}

\AtBeginDocument{
  \abovedisplayskip     =0.5\abovedisplayskip
  \abovedisplayshortskip=0.5\abovedisplayshortskip
  \belowdisplayskip     =0.5\belowdisplayskip
  \belowdisplayshortskip=0.5\belowdisplayshortskip}

\title{Non-uniform Sampling Strategies for NeRF on 360\textdegree~images}

\addauthor{Takashi Otonari}{otonari@hal.t.u-tokyo.ac.jp}{1}
\addauthor{Satoshi Ikehata}{sikehata@nii.ac.jp}{1,2}
\addauthor{Kiyoharu Aizawa}{aizawa@hal.t.u-tokyo.ac.jp}{1}

\addinstitution{
 The University of Tokyo\\
 Tokyo, Japan
}
\addinstitution{
 National Institute of Informatics\\
 Tokyo, Japan
}

\runninghead{Otonari, Ikehata, Aizawa}{Sampling Strategies for NeRF on 360\textdegree~images}

\newcommand{\Fref}[1]{Fig.~\ref{#1}}
\newcommand{\Tref}[1]{Table~\ref{#1}}

\newcommand{\Cref}[1]{Chapter~\ref{#1}}
\newcommand{\eg}[0]{\emph{e.g}.}
\newcommand{\ie}[0]{\emph{i.e}.}

\begin{document}

\maketitle

\begin{abstract}
In recent years, the performance of novel view synthesis using perspective images has dramatically improved with the advent of neural radiance fields (NeRF). This study proposes two novel techniques that effectively build NeRF for 360\textdegree~omnidirectional images. 
Due to the characteristics of a 360\textdegree~image of ERP format that has spatial distortion 
in their high latitude regions and a 360\textdegree~wide viewing angle, NeRF's general ray sampling strategy is ineffective. Hence, the view synthesis accuracy of NeRF is limited and learning is not efficient. 
We propose two non-uniform ray sampling schemes for NeRF to suit 360° images -- distortion-aware ray sampling and content-aware ray sampling.
We created an evaluation dataset \textit{Synth360} using Replica and SceneCity models of indoor and outdoor scenes, respectively. 
In experiments, we show that our proposal successfully builds 360\textdegree~image NeRF in terms of both accuracy and efficiency. 
The proposal is widely applicable to advanced variants of NeRF. DietNeRF, AugNeRF, and NeRF++ combined with the proposed techniques further improve the performance. Moreover, we show that our proposed method enhances the quality of real-world scenes in 360\textdegree~images. 
Synth360: https://drive.google.com/drive/folders/1suL9B7DO2no21ggiIHkH3JF3OecasQLb.
\end{abstract}

\section{Introduction}
Synthesizing a view from other views is a long-standing problem in computer vision and graphics. With recent emerging interest in virtual and augmented reality, this technology is expected to support applications such as virtual tours and immersive 3-D games, where immersion in large, unbounded photorealistic virtual space is possible. In such applications, taking images of an entire scene with a camera of narrow field-of-view (FoV) is prohibitively tedious; therefore, view synthesis with 360\textdegree~cameras of wide FoV is an attractive option. 

Recently, neural radiance fields (NeRF)~\cite{nerf} has brought significant progress in photorealistic novel view synthesis. NeRF is an implicit MLP-based neural network trained on calibrated multi-view images, which maps 5-D vectors (3-D coordinates and 2-D viewing direction) to opacity and color values of the 3-D coordinates viewed from that direction. Using the NeRF model, the image pixels are independently synthesized by accumulating opacity and color values along the camera ray in continuous 3-D space. 

\begin{figure*}[t!]
  \centering
  {\small
  \begin{minipage}[t]{0.27\textwidth}
    \centering
    \includegraphics[height=18mm,width=36mm]{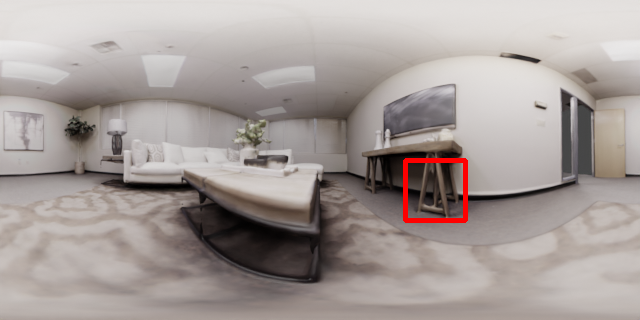}
    \vspace{-0.7cm}\center{Reference}
  \end{minipage}
  \hfill
  \begin{minipage}[t]{0.138\textwidth}
    \centering
    \includegraphics[height=18mm,width=18mm]{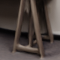}
    \vspace{-0.76cm}\center{\small GT (cropped)}
  \end{minipage}
  \begin{minipage}[t]{0.138\textwidth}
    \centering
    \includegraphics[height=18mm,width=18mm]{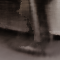}
    \vspace{-0.76cm}\center{\small NeRF~\cite{nerf}\\5,000 iterations}
  \end{minipage}
  \begin{minipage}[t]{0.138\textwidth}
    \centering
    \includegraphics[height=18mm,width=18mm]{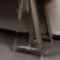}
    \vspace{-0.76cm}\center{\small NeRF~\cite{nerf}\\100,000 iterations}
  \end{minipage}
  \begin{minipage}[t]{0.138\textwidth}
    \centering
    \includegraphics[height=18mm,width=18mm]{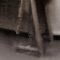}
    \vspace{-0.76cm}\center{\small NeRF + Ours\\5,000 iterations}
  \end{minipage}
  \begin{minipage}[t]{0.138\textwidth}
    \centering
    \includegraphics[height=18mm,width=18mm]{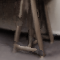}
    \vspace{-0.76cm}\center{\small NeRF + Ours\\100,000 iterations}
  \end{minipage}
  }
\caption{N\"aively sampling rays from wide FoV images causes biased, insufficient training of the NeRF~\cite{nerf} model due to the distorted sphere-to-plane projection and large less-textured regions. Our non-uniform sampling strategies directly tackle this problem.}
\label{fig:tease}
\end{figure*}

Although NeRF~\cite{nerf} and its extensions~(e.g., \cite{tancik2022blocknerf,nerv,yu_and_fridovichkeil2021plenoxels}) have attracted significant attention, to the best of our knowledge, no attempt has been made to train NeRF on images from 360\textdegree~cameras where information at each viewpoint is stored in a single 360\textdegree~image format (e.g., Equirectangular projection (ERP)).

More concretely, the n\"aive NeRF was implemented to uniformly sample rays from each pixel in all training images with equal probability. This is based on the assumption that rays passing through each pixel have equal coverage in 3D space and each pixel in the image has equal information. However, as projecting a spherical 360\textdegree~information of a view to a planar image inevitably introduces projective distortion, the uniform sampling on image coordinates biases the distribution of camera rays in 3-D space. For instance, as shown in~\Fref{fig:tease}, high-latitude areas (\ie, above and below the camera) in an ERP image occupy a much smaller 3-D space compared to low-latitude areas (front and side of the camera). Hence, a strategy is needed to take into consideration the projective distortion of sample rays for training NeRF model on 360\textdegree~images. In addition, a wide FoV image is likely to contain large low-frequency textures (e.g., ceiling and floor, sky and ground) that need smaller samples than high-frequency textures; therefore, n\"aively sampling rays from all the pixels is quite redundant.

The main contribution of this research is to raise these issues involved in applying NeRF to 360\textdegree~images and to present the first effective ideas for addressing them. Concretely, we propose two non-uniform ray sampling strategies for efficiently training NeRF-based models on 360\textdegree~images in the most standardized ERP format. First, we propose the {\it distortion-aware ray sampling}, which normalizes the sampling probability based on the per-pixel scaling factor of the area from the two-dimensional (2-D) plane to the sphere (i.e., the ratio of areas of the spherical surface to those of the corresponding ERP image). Intuitively, pixels at higher latitudes in ERP coordinates have smaller coverage in 3-D space; therefore, the lower sampling probability is allocated. Second, we propose the {\it content-aware ray sampling}, which adaptively updates the sampling probability based on pixel-wise reconstruction loss for each training step. Intuitively, the low-frequency textures such as sky and ground easily converge; therefore, we would lower the sampling probability of those regions as the training proceeds. Our method is surprisingly easy to implement and can therefore be easily incorporated into various NeRF-like models.

Our method is evaluated on both synthetic and real data. To evaluate the proposed method under ideal conditions, we create a synthetic evaluation benchmark of NeRF on 360 images using a physically-based renderer (\ie, {\it Blender}~\cite{blender}) with {\it Replica Dataset}~\cite{replica} and {\it SceneCity}~\cite{scenecity} Blender add-on. In our experiments, we will demonstrate that the proposed non-uniform sampling strategies are applicable to both the n\"aive NeRF~\cite{nerf} and its variants such as DietNeRF~\cite{dietnerf}, AugNeRF~\cite{chen2022augnerf}, and NeRF++~\cite{kaizhang2020}. 

Since most consumer 360\textdegree~cameras store information in an equirectangular projection (ERP) image, we assume that an image is represented in this format as in previous works (\eg, \cite{360_semantic_segmentation,360_depth_estimation,360_super_resolution}). However, it is straightforward to transfer our ideas to other spherical formats, such as cubic projection, which also causes distortions by sphere-to-plane projection. 

\section{Related Works}
\noindent \textbf{Novel View Synthesis}: Before NeRF~\cite{nerf} emerged, discrete representations of scenes have been used in the novel view synthesis task. Several approaches with discrete representations of scenes employ point clouds~\cite{DBLP:journals/corr/abs-1906-08240}, voxels~\cite{sitzmann2019deepvoxels}, meshes~\cite{lettherebecolor}, plane sweep volumes~\cite{deepstereo,deepview}, or multi-plane images~\cite{llff,srinivasan19}. While effective, they have a disadvantage of limited resolution due to the large memory consumption.

Conversely, NeRF~\cite{nerf}, a learning-based method based on continuous implicit representation, has achieved high-resolution rendering by taking advantage of volume rendering with continuous neural radiance fields. To improve the performance of NeRF, several studies have combined multiple representations (e.g., point clouds~\cite{kangle2021dsnerf}, voxels~\cite{liu2020neural}, and multi-plane images~\cite{Wizadwongsa2021NeX,mine2021}). In addition, NeRF opens up many new kinds of research using implicit neural representation~\cite{li2020neural,martinbrualla2020nerfw,nerv2021,Niemeyer2020GIRAFFE}. 

However, there have been few studies on novel view synthesis using multi-view 360\textdegree~images~\cite{DBLP:conf/eccv/LinXMSHDSSR20,DBLP:journals/corr/abs-2103-05842, omninerf}. 
To the best of our knowledge, this work is the first attempt to handle the projective distortion to apply NeRF for calibrated multi-view 360\textdegree~images. Problems that arise when NeRF is applied to 360\textdegree~images in ERP format include projective distortion, the presence of low-frequency texture regions resulting from wide FoV, and unbounded scenes. This study addresses two issues of projective distortion and the presence of regions of low-frequency texture.

The n\"aive NeRF assumes that the entire scene can be packed into a bounded volume, so is problematic for unbounded scenes. NeRF++~\cite{kaizhang2020} and mip-NeRF 360~\cite{barron2022mipnerf360} improved the performance of unbounded scenes in perspective images. NeRF++~\cite{kaizhang2020} separates the scene into foreground and background and parameterizes the background by inverted sphere parameterization. Mip-NeRF 360~\cite{barron2022mipnerf360} proposes a parameterization to handle unbounded scenes under the conical frustum proposed in mip-NeRF~\cite{barron2021mipnerf}, which is a NeRF variant that addresses sampling and aliasing. As a result, the mean squared error is reduced by 54\% compared to mip-NeRF for unbounded scenes in perspective images. However, mip-NeRF 360 is not suitable for images with large projective distortion, such as ERP format. As described by the authors, this is because mip-NeRF assumes a small difference between the base and top radii of the frustum~\cite{barron2021mipnerf}. Our proposed method does not solve the problem caused by this intra-pixel distortion and therefore mip-NeRF 360 is not suitable. Our experiments show that NeRF++ combined with our method makes possible more effective novel view synthesis of 360\textdegree~images for unbounded scenes. 

We should note that some previous works controlled the sampling probability of 3D points along rays (\eg, TermiNeRF~\cite{terminerf} and NeRF-ID~\cite{nerf_in_detail}), however, our proposal is different from them in that we control the {\it pixel-wise} sampling probability.\\

\noindent \textbf{Spherical Novel View Synthesis}: Spherical novel view synthesis is the  task of synthesizing a novel 360\textdegree~view from a set of multi-view 360\textdegree~images. In this field, multi-sphere images (MSI)-based methods, which are spherical extensions of multi-plane images (MPI), are the most commonly used representation~\cite{somsi, Attal:2020:ECCV, broxton2020immersive}. MSI-based methods have a fast rendering time, but provide a discrete representation of the scene and have limited accuracy and large memory consumption to represent large scenes. In contrast, NeRF-based methods have their advantages in their accuracy and practical memory consumption with continuous neural representations even though the large training/test time is still a big challenge. So far, there have been very few attempts to apply NeRF-based models to 360\textdegree~images. If any, OmniNeRF~\cite{omninerf} handled more spherical information from a fisheye camera, and to the best of our knowledge, there was no attempt to explicitly tackle challenges that arise when NeRF-based models are trained from 360\textdegree~information which is projected onto 2-D images. 
\\\\
\noindent \textbf{Hard Example Mining}: In the image understanding tasks such as image classification and object detection, hard example mining is a bootstrapping technique to solve the imbalance problem of training samples~\cite{hem1,hem2}. In hard example mining, the sampling probability is non-uniformly assigned according to its current loss for enhancing the neural network to learn from harder and more important examples. Our content-aware ray sampling strategy introduced the idea of hard example mining in training NeRF model. 

\section{Preliminaries}
Given calibrated multi-view images, NeRF~\cite{nerf} learns implicit 3-D volumes of opacity $\sigma$ and color $\mathbf{c}$ of each 3-D coordinate by minimizing the pixel-wise discrepancy between the actual observation and the volume rendering result. 
The opacity $\sigma$ is a function of 3-D position, independent of the viewing direction, and the color $\mathbf{c}$ is a function of both spatial position and viewing direction. NeRF trains neural networks for both opacity and color volumes based on the inverse volume rendering using regularly sampled 3-D points on rays $\mathbf{r}$ passing through pixels with a ray origin $\mathbf{o}$ and direction $\mathbf{d}$ in each input image as $\mathbf{r}(t) = \mathbf{o} + t\mathbf{d}$ in the range $t_n\leq t\leq t_f$. Specifically, we partition $[t_n, t_f]$ into $N$ evenly-spaced bins and then draw one sample $t_i (1 \leq i\leq N)$ uniformly at random from within each bin as
\begin{align}
    &t_i \sim \mathcal{U}\left[t_n+\frac{i-1}{N}(t_f-t_n), t_n+\frac{i}{N}(t_f-t_n)\right].
\end{align}
Using all samples along a ray, the color $\hat{C}(\mathbf{r})$ at a corresponding pixel is computed using the volume rendering principle as
\begin{align}
    \hat{C}(\mathbf{r})=&\sum_{i=1}^{N}T_i(1-\exp(-\sigma_i\delta_i))\mathbf{c}_i, \mathrm{where}\ T_i = \exp\left(-\sum_{j=1}^{i-1}\sigma_j\delta_j\right),
    \label{eq:NeRF_volume_rendering}
\end{align}
where $\delta_i=t_{i+1}-t_i$ is the distance between adjacent samples. NeRF has two levels of coarse and fine MLPs and learns for the loss function $\mathcal{L}$, that is, the squared error between the color $\hat{C}_c(\mathbf{r})$, $\hat{C}_f(\mathbf{r})$ synthesized by each ray $\mathbf{r}\in\mathcal{R}$ in each batch and the ground truth color $C(\mathbf{r})$.
\begin{align}
    \mathcal{L}=\sum_{\mathbf{r}\in\mathcal{R}} \bigg\lbrack\Big\lVert \hat{C}_c(\mathbf{r})-C(\mathbf{r})\Big\rVert_2^2+\Big\lVert \hat{C}_f(\mathbf{r})-C(\mathbf{r})\Big\rVert_2^2\bigg\rbrack. \label{eq:NeRF_loss_function}
\end{align}

\section{Method}
Basically, NeRF is trained on the batch-wise reconstruction loss where each batch contains multiple rays whose sampling probability is uniform at all the pixels in all the images. However, as has already been stated, when NeRF model is trained on 360\textdegree~images in ERP format, this uniform sampling strategy theoretically becomes problematic as the 3-D coverage of each ray passing through each pixel is {\it not} uniform due to the projective distortion. In addition, a wide FoV image basically contains large low-frequency texture regions such as sky and ground, ceiling and floor, and it is wasteful to keep spending the same amount of learning resources there as are spent on the high-textured areas. Therefore, we propose two {\it non-uniform} ray sampling strategies that individually control the pixel-wise sampling probability considering the geometric distortion (\ie, distortion-aware ray sampling) and image content (\ie, content-aware ray sampling). 
\begin{figure}[!t]
\centering
\begin{minipage}[b]{0.51\linewidth}
  \centering
  \centerline{\includegraphics[width=\linewidth]{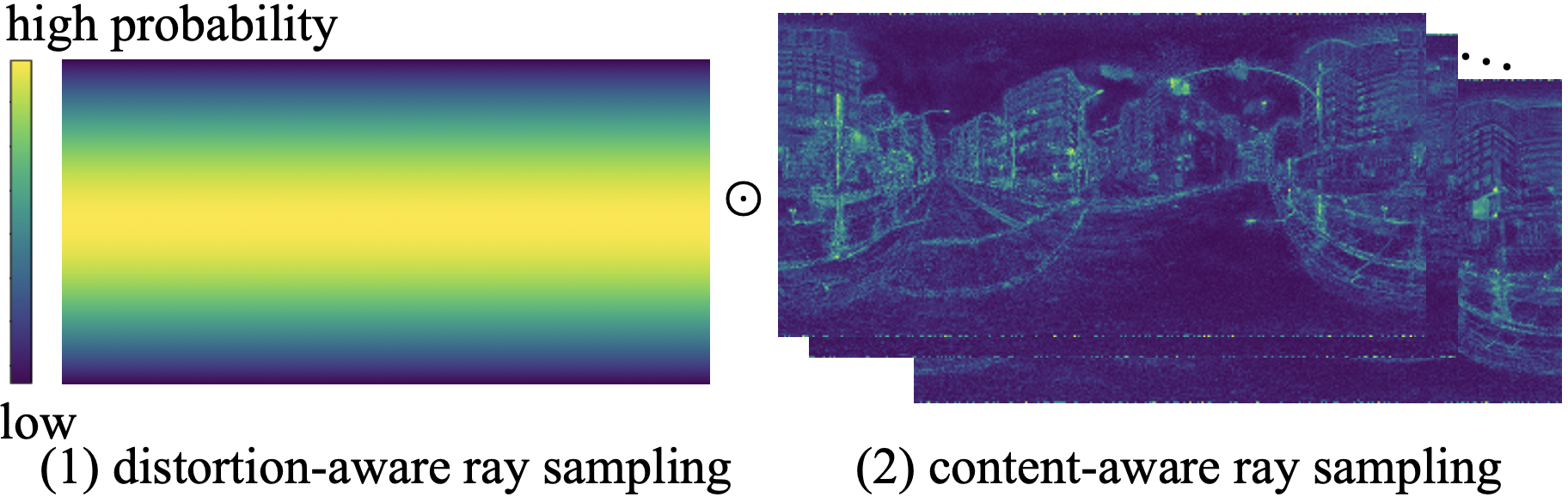}}
  \centerline{(a) two non-uniform sampling strategies}
\end{minipage}
\begin{minipage}[b]{0.48\linewidth}
    \begin{minipage}[b]{0.32\linewidth}
      \centering
      \centerline{\includegraphics[width=\linewidth]{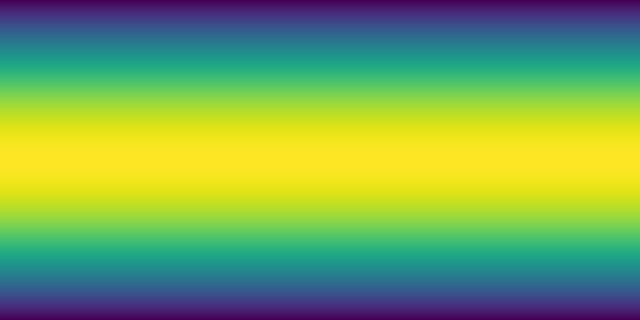}}
      \centerline{0 iterations}
    \end{minipage}
    \begin{minipage}[b]{0.32\linewidth}
      \centering
      \centerline{\includegraphics[width=\linewidth]{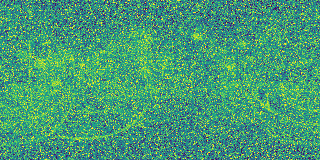}}
      \centerline{500}
    \end{minipage}
    \begin{minipage}[b]{0.32\linewidth}
      \centering
      \centerline{\includegraphics[width=\linewidth]{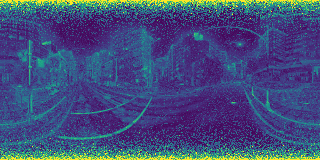}}
      \centerline{1,000}
    \end{minipage}
    \begin{minipage}[b]{0.32\linewidth}
      \centering
      \centerline{\includegraphics[width=\linewidth]{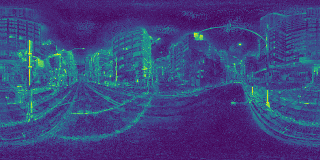}}
      \centerline{2,500}
    \end{minipage}
    \begin{minipage}[b]{0.32\linewidth}
      \centering
      \centerline{\includegraphics[width=\linewidth]{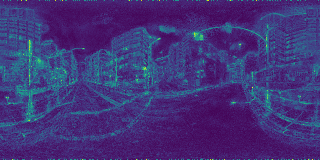}}
      \centerline{10,000}
    \end{minipage}
    \begin{minipage}[b]{0.32\linewidth}
      \centering
      \centerline{\includegraphics[width=\linewidth]{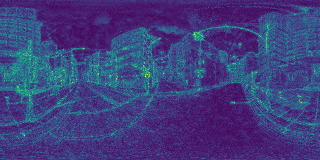}}
      \centerline{100,000}
    \end{minipage}
  \centerline{(b) transition in sampling probability}
\end{minipage}
\caption{Ray sampling of the proposed method. The proposed method is divided into two components: (1) distortion-aware ray sampling and (2) content-aware ray sampling.}
\label{fig:proposed}
\end{figure}
\\\\
\noindent \textbf{Distortion-aware Ray Sampling}:
To reduce the sphere-to-plane projective distortion bias by uniform sampling of rays for the inverse volume rendering in NeRF~\cite{nerf}, we control the sampling probability pixel-by-pixel considering the coverage of each pixel in 3-D space. Concretely, we compute the area $S_d$ where a pixel of an ERP image occupies the unit spherical surface as
\begin{align}
    S_d &= \int_{\phi_1}^{\phi_2}\int_{\theta_1}^{\theta_2}\cos\theta d\theta d\phi \notag \\
    &= (\phi_2-\phi_1)(\sin\theta_2-\sin\theta_1). \label{eq:solid_angle}
\end{align}
Here $\theta\in[\theta_1, \theta_2]$ and $\phi\in[\phi_1, \phi_2]$ are latitude and longitude respectively, that bound each pixel in the ERP coordinates. Simply speaking, the higher latitude regions have smaller 3-D coverage and lower latitude regions have larger 3-D coverage. We calculate $S_d$ for all pixels in all training images and normalize all of the values so that they add up to one. This result is used as the probability of sampling each pixel ($P_d$) during training as shown in~\Fref{fig:proposed}-(a)-left. Intuitively, the higher sampling probabilities are assigned to lower latitude regions as each ray has to cover the larger space in 3-D.
\\\\
\noindent \textbf{Content-aware Ray Sampling}:
To avoid taking redundant samples from low-frequency texture (texture-less) regions in a 360\textdegree~image with wide FoV, we further control the sampling probability so that the probability around the texture-less regions decreases. However, if no samples were assigned to the low-frequency region at all, learning would not proceed in that region; therefore, it is desirable to take samples from the entire image in the early stages of learning, then samples gradually be concentrated in more challenging regions. Based on this observation, we are inspired by the online hard example mining~\cite{hem1,hem2}, which is a bootstrapping technique that adaptively samples examples in a non-uniform way depending on the current loss of each example. We assume that the $\ell_2$ reconstruction loss around the low-frequency texture regions decreases faster than that around the high-frequency texture regions; hence, we decrease the sampling probability at pixels with smaller reconstruction loss at the {\it last} iteration. Concretely, we define $S_c$ as a collection of pixel-wise inverse reconstruction loss at all pixels in all the training images that are uniformly initialized by one. At each iteration, a batch of rays passing through $m$ pixels  (\ie, $m=2048$ in our implementation) are stochastically sampled using the sampling probability at the current iteration and only $S_c$ values at sampled pixels are updated based on the reconstruction loss. Then, the content-aware sampling probability ($P_c$) of each pixel is updated  by normalizing $S_c$ so that they add up to one (See~\Fref{fig:proposed}-(a)-right). In~\Fref{fig:proposed}-(b), we illustrate how the content-aware sampling probabilities are updated through iterations.
\\\\
\noindent\textbf{Sampling Strategy Details}: We multiply and normalize the distortion-aware probabilities ($P_d$) and content-aware probabilities ($P_c$) so that they sum up to one. At each training iteration, we stochastically pick $m$ pixels using this sampling probability, and a NeRF model is trained on the rays that pass through those pixels. We perturb the center of rays within the pixels to augment their coverage.


\section{Results}
To validate the effectiveness of our non-uniform ray sampling strategies, we implemented our method on NeRF~\cite{nerf} and its variants (\ie, DietNeRF~\cite{dietnerf}, AugNeRF~\cite{chen2022augnerf}, and NeRF++~\cite{kaizhang2020}), and trained each model on both synthetic and real datasets. To exclude the effects other than different sampling strategies, only the minimal changes for the adaptation to ERP images were made (\ie, a ray is defined on the spherical coordinates, rather than Cartesian coordinates). All the models were trained on a single NVIDIA Tesla A100 machine with 2048 samples per iteration. We used the Adam optimizer with default hyperparameters  (\ie, $\beta_1=$0.9, $\beta_2=$0.999, and $\epsilon=10^{-7}$) and a learning rate of $5\times10^{-4}$ which was linearly decayed so that it became $5\times10^{-5}$ at the 100k-th iteration. For each target scene, models were trained during 100k iterations taking approximately 7$\sim$8 hours. In all scenes, the ray's near values were set to 0. In synthetic scenes, the ray's far values were set to reach the area excluding the sky. In real-world scenes, sufficient far value to reach the point clouds was set based on the OpenSfM camera position and point clouds.

\subsection{Synth360 Dataset}
\begin{figure}[!t]
  \centering
  \centerline{\includegraphics[width=\linewidth]{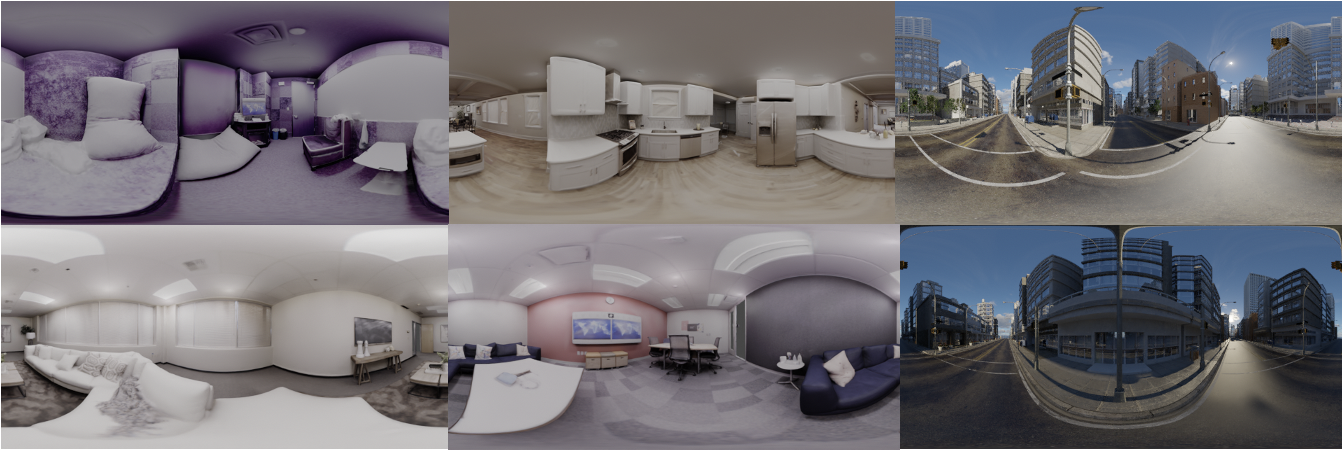}}
  \vspace{-2mm}
  \caption{Examples of synthesized scenes (\ie, 6 of 10 scenes) in Synth360 dataset.}
  \label{fig:eval_dataset}
\end{figure}

It is known that the evaluation of NeRF-based models using real-world datasets is inevitably affected by camera calibration errors by structure-from-motion (SfM) tools, which may negatively affect the theoretical analysis~\cite{nerfmm}. Furthermore, a real 360\textdegree~image often suffers from stitching errors which destroy the precise geometric consistency. Thus, we firstly evaluated our method on the ideal synthetic images. Since there was little synthetic dataset available for novel-view synthesis with multiple 360\textdegree~images, we synthesized 360\textdegree~images in the ERP format with Blender's Cycles renderer~\cite{blender} using highly photorealistic 3D scenes, Replica Dataset~\cite{replica}, for indoor scenes and the city generator add-on for Blender, SceneCity~\cite{scenecity}, for outdoor scenes. Using these resources, we randomly picked 8 indoor scenes from Replica Dataset and generated 2 city scenes with SceneCity. For each indoor and outdoor scene, we rendered images by placing 5 to 9 cameras uniformly for training images and 100 cameras for test images whose resolutions are all $320\times 640$. Examples of rendered images are illustrated in~\Fref{fig:eval_dataset} 

\begin{figure}[!t]
\begin{minipage}[b]{0.49\linewidth}
    \begin{minipage}[b]{0.49\linewidth}
      \centering
      \centerline{\includegraphics[width=\linewidth]{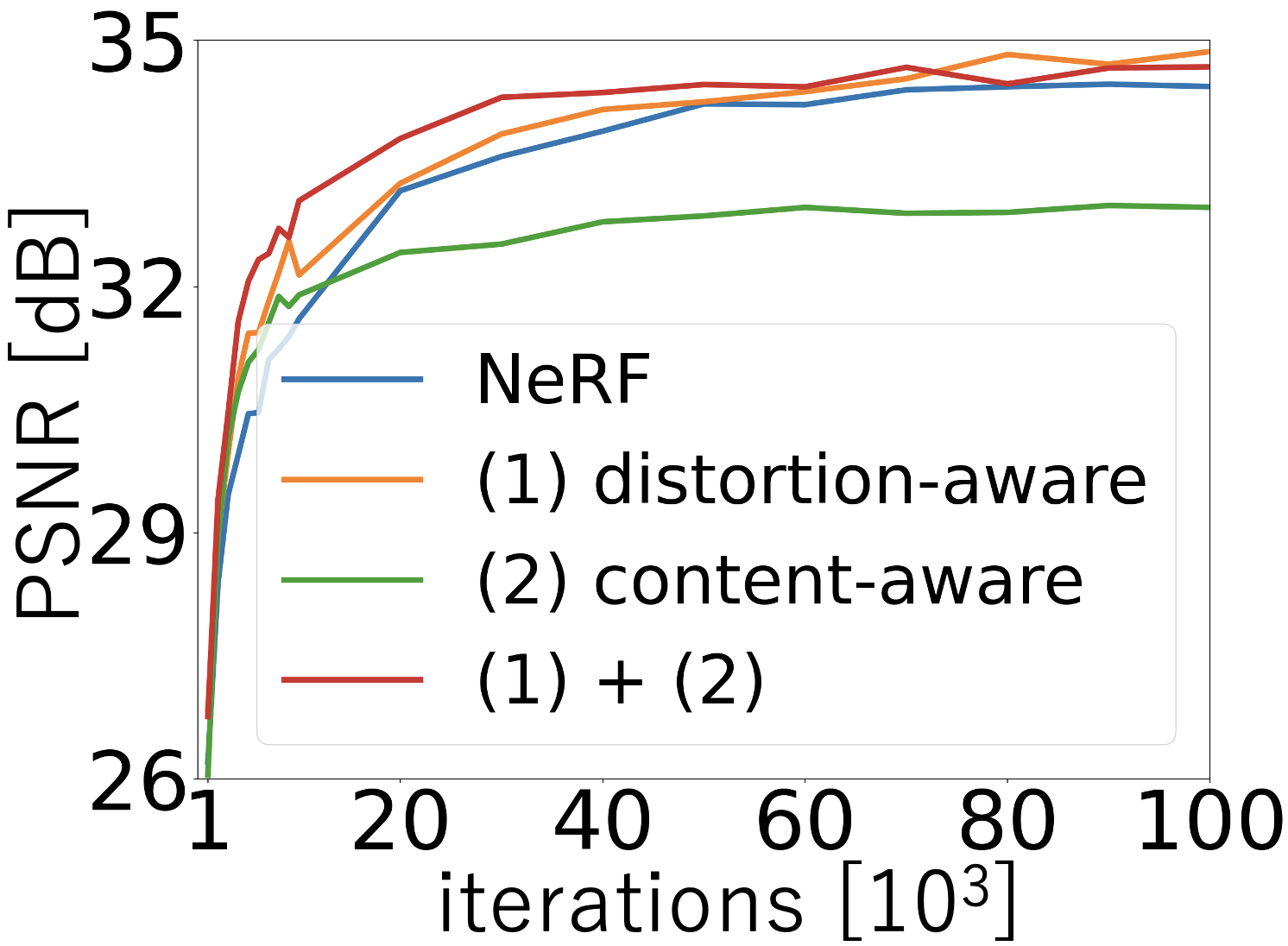}}
    \end{minipage}
    \hfill
    \begin{minipage}[b]{0.49\linewidth} 
      \centering
      \centerline{\includegraphics[width=\linewidth]{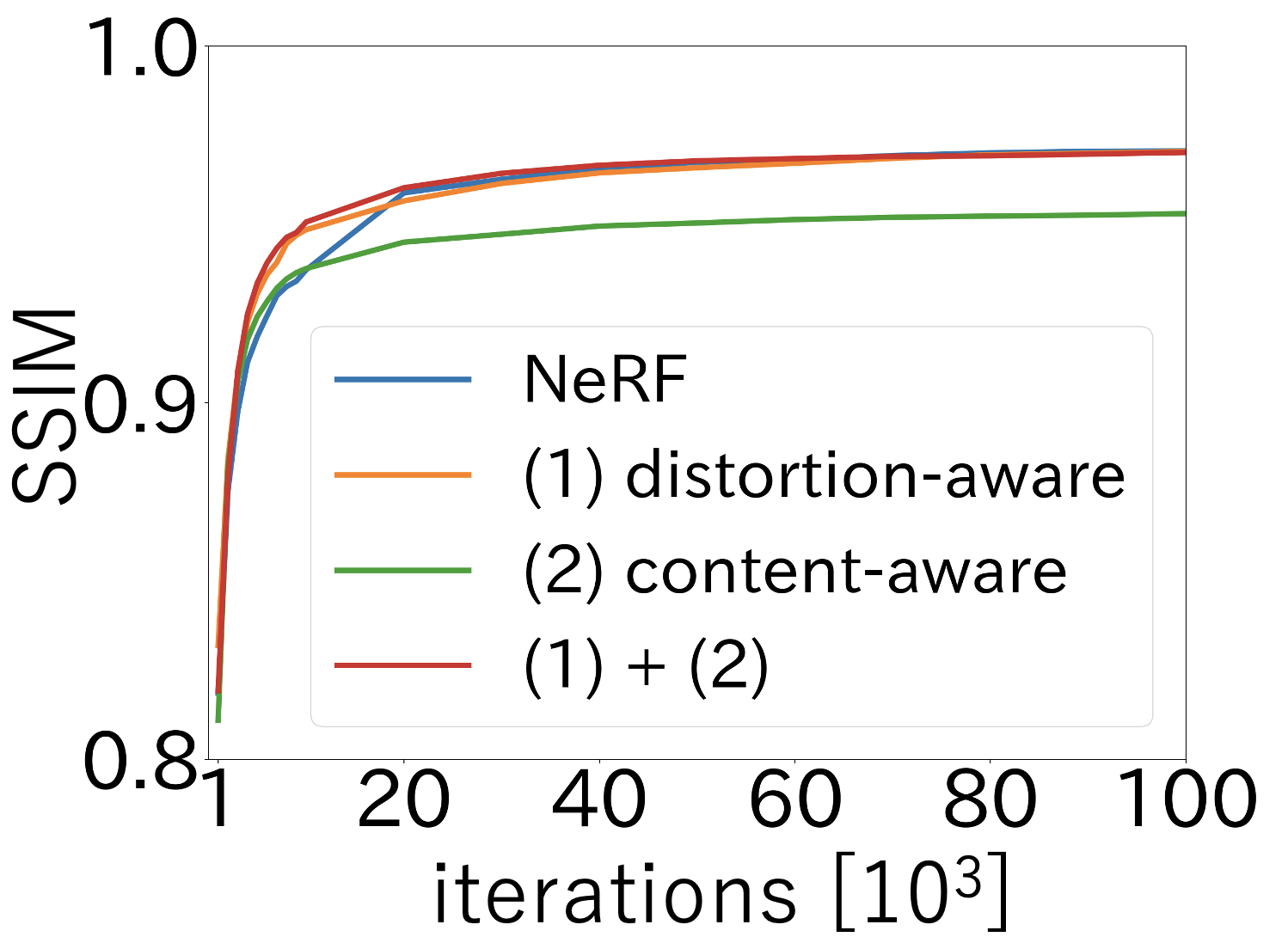}}
    \end{minipage}
    \centerline{Indoor scenes (from Replica Dataset~\cite{replica})}
\end{minipage}
\begin{minipage}[b]{0.49\linewidth}
    \begin{minipage}[b]{0.49\linewidth}
      \centering
      \centerline{\includegraphics[width=\linewidth]{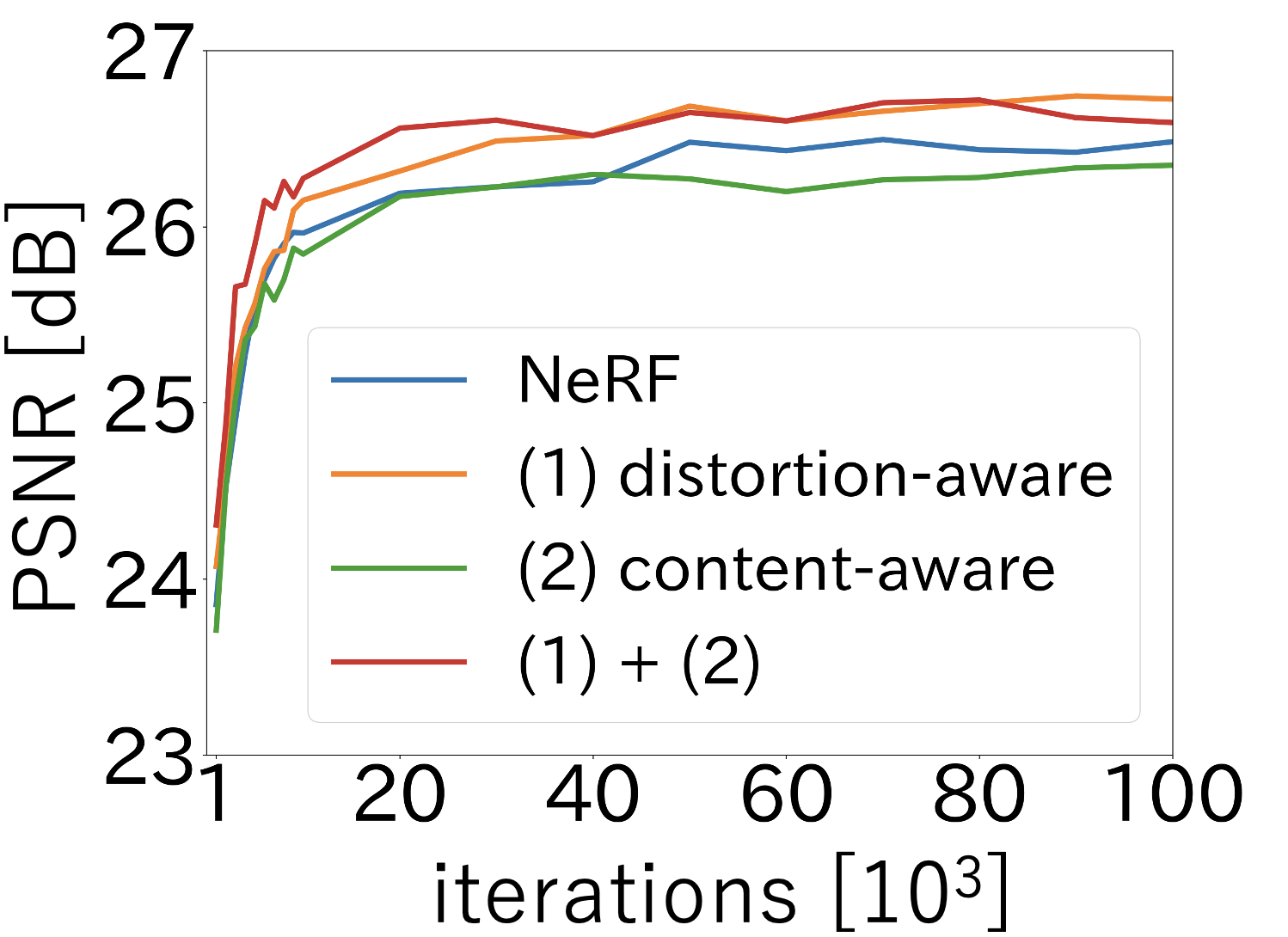}}
    \end{minipage}
    \hfill
    \begin{minipage}[b]{0.49\linewidth} 
      \centering
      \centerline{\includegraphics[width=\linewidth]{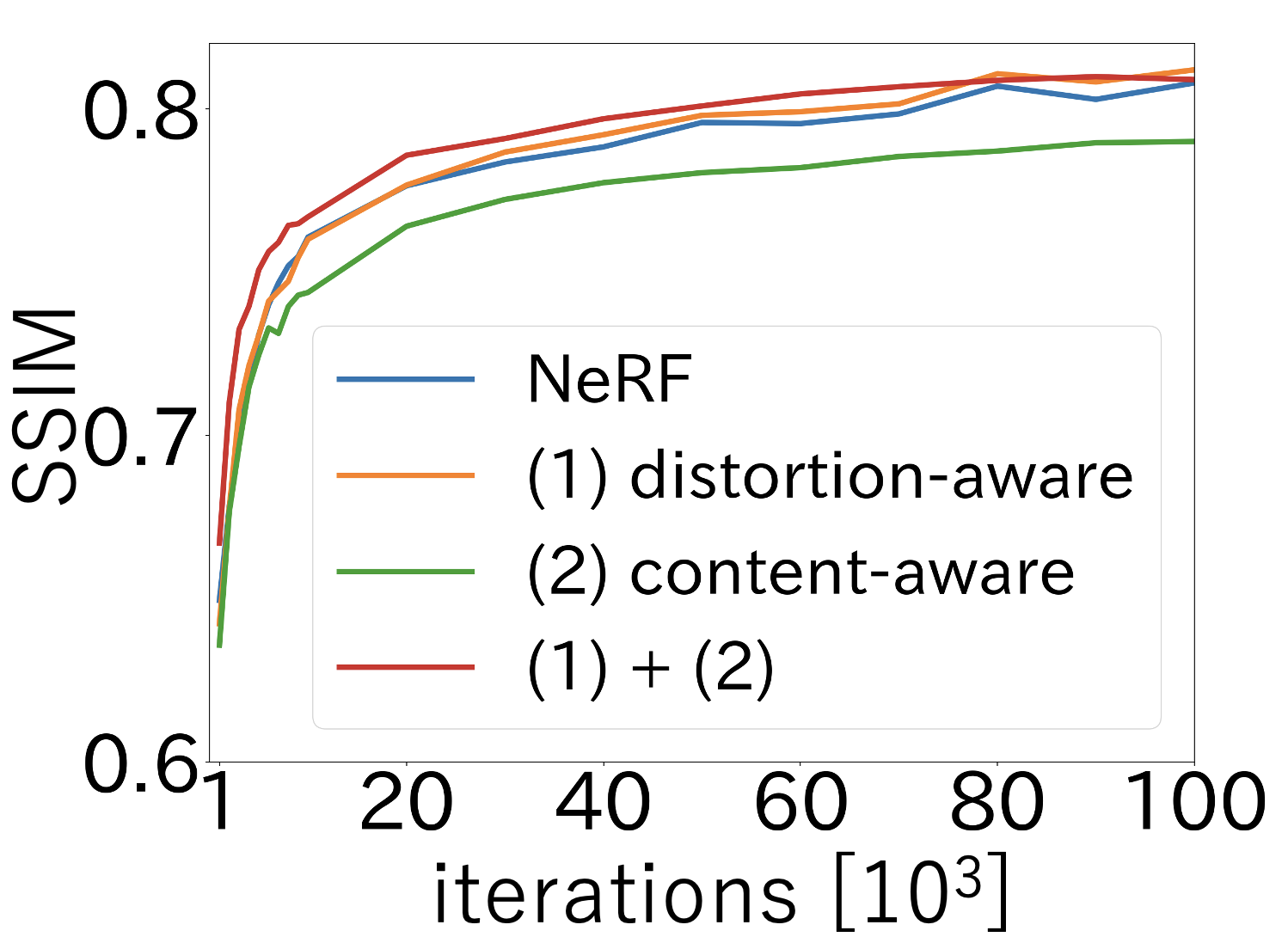}}
    \end{minipage}
    \centerline{Outdoor scenes (from SceneCity~\cite{scenecity})}
\end{minipage}
\caption{Effect of each sampling strategy. The proposed method is divided into two parts: (1) distortion-aware ray sampling and (2) content-aware ray sampling. PSNR/SSIM were calculated for the images at each test pose.}
\label{fig:experiment_psnr_graph}
\end{figure}

\begin{figure*}[t!]
  \centering
  {\small
  \begin{minipage}[t]{0.27\textwidth}
    \centering
    \includegraphics[height=18mm,width=36mm]{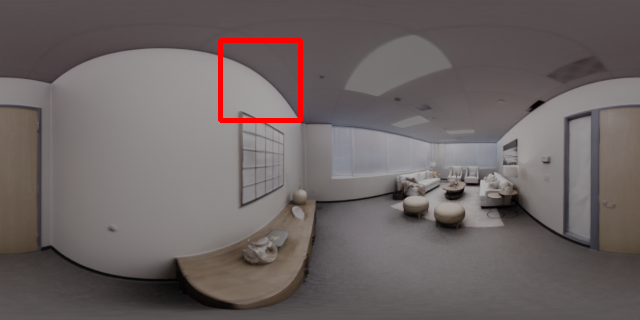}
    \vspace{-0.7cm}\center{Replica Dataset~\cite{replica}}
  \end{minipage}
  \hfill
  \begin{minipage}[t]{0.138\textwidth}
    \centering
    \includegraphics[height=18mm,width=18mm]{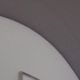}
    \vspace{-0.76cm}\center{\small GT (cropped)}
  \end{minipage}
  \begin{minipage}[t]{0.138\textwidth}
    \centering
    \includegraphics[height=18mm,width=18mm]{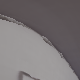}
    \vspace{-0.76cm}\center{\small NeRF~\cite{nerf}\\5,000 iterations}
  \end{minipage}
  \begin{minipage}[t]{0.138\textwidth}
    \centering
    \includegraphics[height=18mm,width=18mm]{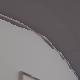}
    \vspace{-0.76cm}\center{\small NeRF~\cite{nerf}\\100,000 iterations}
  \end{minipage}
  \begin{minipage}[t]{0.138\textwidth}
    \centering
    \includegraphics[height=18mm,width=18mm]{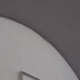}
    \vspace{-0.76cm}\center{\small NeRF + Ours\\5,000 iterations}
  \end{minipage}
  \begin{minipage}[t]{0.138\textwidth}
    \centering
    \includegraphics[height=18mm,width=18mm]{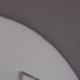}
    \vspace{-0.76cm}\center{\small NeRF + Ours\\100,000 iterations}
  \end{minipage}
  \\
  \begin{minipage}[t]{0.27\textwidth}
    \centering
    \includegraphics[height=17.8mm,width=35.6mm]{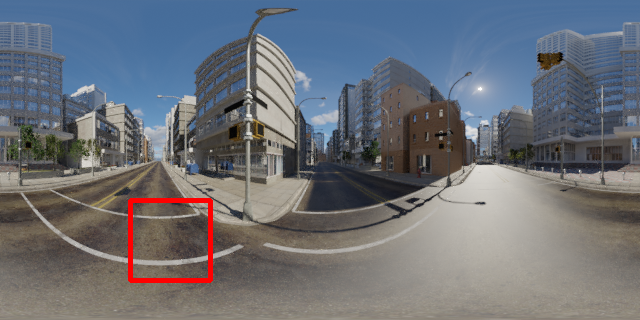}
    \vspace{-0.7cm}\center{SceneCity~\cite{scenecity}}
  \end{minipage}
  \hfill
  \begin{minipage}[t]{0.138\textwidth}
    \centering
    \includegraphics[height=18mm,width=18mm]{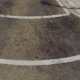}
    \vspace{-0.76cm}\center{\small GT (cropped)}
  \end{minipage}
  \begin{minipage}[t]{0.138\textwidth}
    \centering
    \includegraphics[height=18mm,width=18mm]{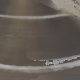}
    \vspace{-0.76cm}\center{\small NeRF~\cite{nerf}\\5,000 iterations}
  \end{minipage}
  \begin{minipage}[t]{0.138\textwidth}
    \centering
    \includegraphics[height=18mm,width=18mm]{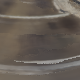}
    \vspace{-0.76cm}\center{\small NeRF~\cite{nerf}\\100,000 iterations}
  \end{minipage}
  \begin{minipage}[t]{0.138\textwidth}
    \centering
    \includegraphics[height=18mm,width=18mm]{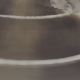}
    \vspace{-0.76cm}\center{\small NeRF + Ours\\5,000 iterations}
  \end{minipage}
  \begin{minipage}[t]{0.138\textwidth}
    \centering
    \includegraphics[height=18mm,width=18mm]{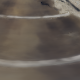}
    \vspace{-0.76cm}\center{\small NeRF + Ours\\100,000 iterations}
  \end{minipage}\\
  }
\caption{Qualitative comparison on Replica Dataset~\cite{replica} and SceneCity~\cite{scenecity}.}
\label{fig:experiment1_image}
\end{figure*}

\subsection{Evaluation on Synth360}
For evaluating the contribution of individual components, we firstly ablated each of our non-uniform sampling strategies implemented upon the n\"aive NeRF model~\cite{nerf}. PSNR/SSIM curves during training are shown in~\Fref{fig:experiment_psnr_graph}. By comparing curves between the n\"aive NeRF model and that with both distortion-aware and content-aware sampling strategies, we observe the obvious performance improvement by our method on both indoor (Replica) and outdoor (SceneCity) scenes; not only did our method improved the learning speed, but also PSNR/SSIM at convergence. \Fref{fig:experiment1_image} illustrates the qualitative comparisons at 5,000-th iteration and 100,000-th iteration, which show obvious advantages of our method over n\"aive NeRF to recover details around the boundaries between walls and ceilings~(\Fref{fig:experiment1_image}-top) and around the white centerline~(\Fref{fig:experiment1_image}-bottom). Looking at the results in~\Fref{fig:experiment_psnr_graph} in more details, we also observe that the content-aware sampling alone rather degraded the performance. We further analyzed this failure case and found that the model trained based solely on the content-aware sampling without the distortion-aware sampling had been overfitted to high-frequency details at high-latitude regions which were severely distorted due to the projective distortion. This result suggests that a proper consideration of the projective distortion is critical for NeRF on 360\textdegree~images. In the supplementary material, we provide a deeper analysis of how our distortion-aware sampling contributed to different latitude regions and how our content-aware sampling contributed to a different amount of textures. 

To validate that the benefit of our non-uniform sampling strategies is not limited to the n\"aive NeRF model, we applied our method to other NeRF-like models such as DietNeRF~\cite{dietnerf}, AugNeRF~\cite{chen2022augnerf}, and NeRF++~\cite{kaizhang2020} with only changes about sampling strategies.\footnote{More implementation details are presented in the supplementary.} A quantitative comparison of PSNR/SSIM at 100,000-th iteration is shown in~\Tref{table:experiment3_dietnerf}. We observe that our non-uniform sampling strategy consistently improved the performance of DietNeRF, AugNeRF, and NeRF++. Please note that the reconstruction accuracy of NeRF++ is much lower than others because NeRF++ failed to properly decompose the foreground and background on our Synth360 dataset. In the supplementary, we also visualize PSNR/SSIM curves and rendered images for each method for further discussion. 

\begin{table}[!t]
    \centering
    \caption{Quantitative comparison using NeRF~\cite{nerf}, DietNeRF~\cite{dietnerf}, AugNeRF~\cite{chen2022augnerf}, and NeRF++~\cite{kaizhang2020} on Replica Dataset~\cite{replica} and SceneCity~\cite{scenecity}. The best is highlighted.}
    \label{table:experiment3_dietnerf}
    \begin{tabular}{c|rr|rr}
    & \multicolumn{2}{|c|}{Replica Dataset~\cite{replica}} &  \multicolumn{2}{c}{SceneCity~\cite{scenecity}}\\\hline
    Method & PSNR$\uparrow$ & SSIM$\uparrow$ & PSNR$\uparrow$ & SSIM$\uparrow$ \\\hline
    NeRF~\cite{nerf} & 34.44  & 0.970 & 26.48  & 0.808 \\
    NeRF~\cite{nerf}+Ours & 34.68  & 0.970 & 26.59  & 0.809 \\
    DietNeRF~\cite{dietnerf} & 36.34  & 0.974 & 31.86  & 0.889 \\
    DietNeRF~\cite{dietnerf}+Ours & \textbf{37.69} & \textbf{0.977} & \textbf{32.26} & \textbf{0.892} \\
    AugNeRF~\cite{chen2022augnerf} & 36.80  & 0.982 & 27.53  & 0.826 \\
    AugNeRF~\cite{chen2022augnerf}+Ours & 37.07 & 0.983 & 28.61 & 0.844 \\
    NeRF++~\cite{kaizhang2020} & 12.39  & 0.386 & 26.62  & 0.799 \\
    NeRF++~\cite{kaizhang2020}+Ours & 14.39 & 0.521 & 28.33 & 0.832 \\
    \end{tabular}
\end{table}

\subsection{Evaluation on Real 360\textdegree~Images}
\begin{figure*}[t!]
  \centering
  {\small
  \begin{minipage}[t]{0.27\textwidth}
    \centering
    \includegraphics[height=18mm,width=36mm]{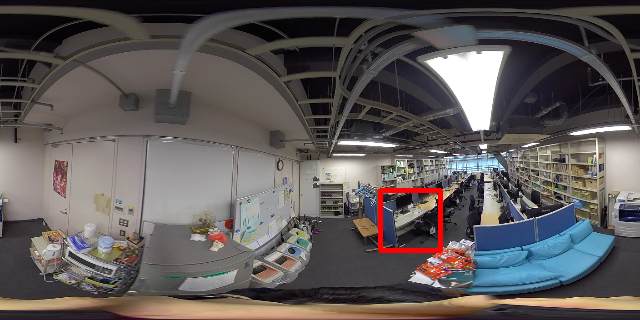}
    \vspace{-0.7cm}\center{indoor scene}
  \end{minipage}
  \hfill
  \begin{minipage}[t]{0.138\textwidth}
    \centering
    \includegraphics[height=18mm,width=18mm]{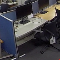}
    \vspace{-0.76cm}\center{\small GT (cropped)}
  \end{minipage}
  \begin{minipage}[t]{0.138\textwidth}
    \centering
    \includegraphics[height=18mm,width=18mm]{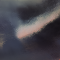}
    \vspace{-0.76cm}\center{\small NeRF~\cite{nerf}\\5,000 iterations}
  \end{minipage}
  \begin{minipage}[t]{0.138\textwidth}
    \centering
    \includegraphics[height=18mm,width=18mm]{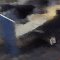}
    \vspace{-0.76cm}\center{\small NeRF~\cite{nerf}\\100,000 iterations}
  \end{minipage}
  \begin{minipage}[t]{0.138\textwidth}
    \centering
    \includegraphics[height=18mm,width=18mm]{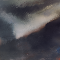}
    \vspace{-0.76cm}\center{\small NeRF + Ours\\5,000 iterations}
  \end{minipage}
  \begin{minipage}[t]{0.138\textwidth}
    \centering
    \includegraphics[height=18mm,width=18mm]{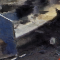}
    \vspace{-0.76cm}\center{\small NeRF + Ours\\100,000 iterations}
  \end{minipage}
  \\
  \begin{minipage}[t]{0.27\textwidth}
    \centering
    \includegraphics[height=18mm,width=36mm]{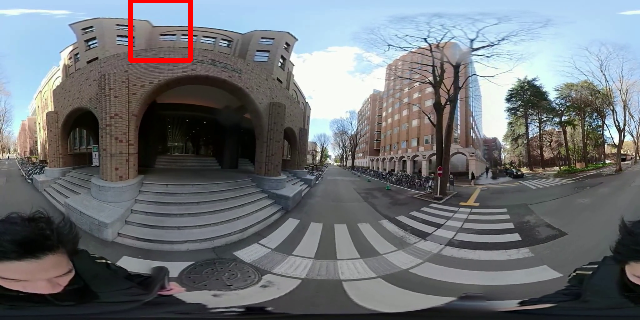}
    \vspace{-0.7cm}\center{outdoor scene}
  \end{minipage}
  \hfill
  \begin{minipage}[t]{0.138\textwidth}
    \centering
    \includegraphics[height=18mm,width=18mm]{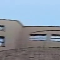}
    \vspace{-0.76cm}\center{\small GT (cropped)}
  \end{minipage}
  \begin{minipage}[t]{0.138\textwidth}
    \centering
    \includegraphics[height=18mm,width=18mm]{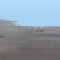}
    \vspace{-0.76cm}\center{\small NeRF~\cite{nerf}\\5,000 iterations}
  \end{minipage}
  \begin{minipage}[t]{0.138\textwidth}
    \centering
    \includegraphics[height=18mm,width=18mm]{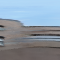}
    \vspace{-0.76cm}\center{\small NeRF~\cite{nerf}\\100,000 iterations}
  \end{minipage}
  \begin{minipage}[t]{0.138\textwidth}
    \centering
    \includegraphics[height=18mm,width=18mm]{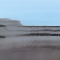}
    \vspace{-0.76cm}\center{\small NeRF + Ours\\5,000 iterations}
  \end{minipage}
  \begin{minipage}[t]{0.138\textwidth}
    \centering
    \includegraphics[height=18mm,width=18mm]{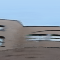}
    \vspace{-0.76cm}\center{\small NeRF + Ours\\100,000 iterations}
  \end{minipage}
  }
  \caption{Qualitative comparison on our real-world indoor and outdoor scenes.}
\label{fig:realworld_image_main}
 \end{figure*}

We also evaluated our non-uniform sampling strategies on two real scenes; one is indoor and the other is outdoor. We used a consumer 360\textdegree~camera to capture each scene in the ERP format and calibrated extrinsic parameters using OpenSfM~\cite{opensfm}. A quantitative comparison of PSNR/SSIM at 100,000-th iteration among ours (with both strategies), NeRF~\cite{nerf}, DietNeRF~\cite{dietnerf}, AugNeRF~\cite{chen2022augnerf} and NeRF++~\cite{kaizhang2020} on these two scenes is shown in~\Tref{table:experiment3_realworld}. While PSNR/SSIM scores of real images are lower than those of synthetic ones as expected, our proposed method also consistently improved the reconstruction quality. We illustrate the qualitative comparison at 5,000-th iteration and 100,000-th iteration in~\Fref{fig:realworld_image_main}.\footnote{Due to the space limit, we only show the rendered images of NeRF and NeRF + Ours. Please refer to the supplementary for other results.} It is interesting to see that our sampling strategies did not significantly improve both PSNR and SSIM scores from the n\"aive NeRF, however, we can observe the clear advantages of our method on the visual comparison. In 360\textdegree~images with a wide field of view and many flat areas, non-uniform sampling seemed to contribute to the visual quality of high-frequency regions more than PSNR/SSIM scores. 

\begin{table}[!t]
    \centering
    \caption{Quantitative comparison using NeRF~\cite{nerf}, DietNeRF~\cite{dietnerf}, AugNeRF~\cite{chen2022augnerf}, and NeRF++~\cite{kaizhang2020} on our real-world indoor and outdoor scenes. The best is highlighted.}
    \label{table:experiment3_realworld}
    \begin{tabular}{c|rr|rr}
    & \multicolumn{2}{|c|}{indoor scene} &  \multicolumn{2}{c}{outdoor scene}\\\hline
    Method & PSNR$\uparrow$ & SSIM$\uparrow$ & PSNR$\uparrow$ & SSIM$\uparrow$ \\\hline
    NeRF~\cite{nerf} & 22.84 & 0.760 & 24.14 & 0.723 \\
    NeRF~\cite{nerf}+Ours & 23.28 & \textbf{0.805} & 24.16 & 0.739 \\
    DietNeRF~\cite{dietnerf} & 21.50 & 0.771 & 23.58 & 0.733 \\
    DietNeRF~\cite{dietnerf}+Ours & 22.89 & 0.794 & 23.84 & 0.735 \\
    AugNeRF~\cite{chen2022augnerf} & 18.71 & 0.677 & 22.44 & 0.682 \\
    AugNeRF~\cite{chen2022augnerf}+Ours & 20.60 & 0.686 & 22.54 & 0.724 \\
    NeRF++~\cite{kaizhang2020} & 22.20 & 0.769 & 23.90 & 0.786\\
    NeRF++~\cite{kaizhang2020}+Ours & \textbf{23.37} & 0.801 & \textbf{24.32} & \textbf{0.803} \\
    \end{tabular}
\end{table}

\section{Conclusion}
In this work, we proposed two non-uniform ray sampling strategies for effectively training NeRF-based models from 360\textdegree~images specifically in the ERP format: distortion-aware ray sampling and content-aware ray sampling. Based on our Synth360 dataset which rendered synthetic indoor and outdoor scenes, we showed that the proposed method consistently improved the training curves of n\"aive NeRF and its variants. Because of its simplicity, the proposed method is highly compatible with other efficient methods (\eg,~\cite{mueller2022instant}), and we believe that the most important contribution of this research is that it shows the importance of non-uniform sampling of rays in distorted image representations, such as 360\textdegree~images. For future work, we should integrate our non-uniform sampling strategies into more diverse variants of NeRF model (\eg, ~\cite{li2020neural,martinbrualla2020nerfw,nerv2021,Niemeyer2020GIRAFFE}).

\section*{Acknowledgement}
This work is partially supported by JSPS KAKENHI 21H03460 and JST-Mirai Program JPMJMI21H1.

\bibliography{egbib}
\end{document}


\maketitle

\section{More Results with DietNeRF, AugNeRF and NeRF++ on Synth360}
\subsection{Details of Methods}
In the main paper, we demonstrated the effectiveness of our non-uniform sampling strategies in different NeRF-like models. We have implemented the proposed method in the authors' publicly available source codes with minimal changes. Here we briefly describe the details of individual algorithms.

DietNeRF~\cite{dietnerf} added an auxiliary semantic consistency loss to the n\"aive NeRF. The weight of an auxiliary semantic consistency loss was set to 0.01 and calculated by using an image that was resized to 224$\times$224 for every 10 iterations. 

AugNeRF~\cite{chen2022augnerf} brought the power of robust data augmentations into regularizing the NeRF training. Augmentation was given to the intermediate feature, pre-rendering output, and input coordinate levels. The perturbations were estimated by multi-step Projected Gradient Descent. The weights of photometric loss and adversarial loss were set to 0.5.

NeRF++~\cite{kaizhang2020} was specially optimized for unbounded scenes by explicitly decomposing an entire scene into the foreground and background ones with different distance divisions. For this decomposition, NeRF++ normalizes each scene so that all cameras were inside the sphere of radius $\frac{1}{1.2}$ and the average camera position of all the training cameras was the center of the sphere.

\subsection{PSNR/SSIM Curves}
\begin{figure}[!t]
\begin{minipage}[b]{0.49\linewidth}
    \begin{minipage}[b]{0.49\linewidth}
      \centering
      \centerline{\includegraphics[width=\linewidth]{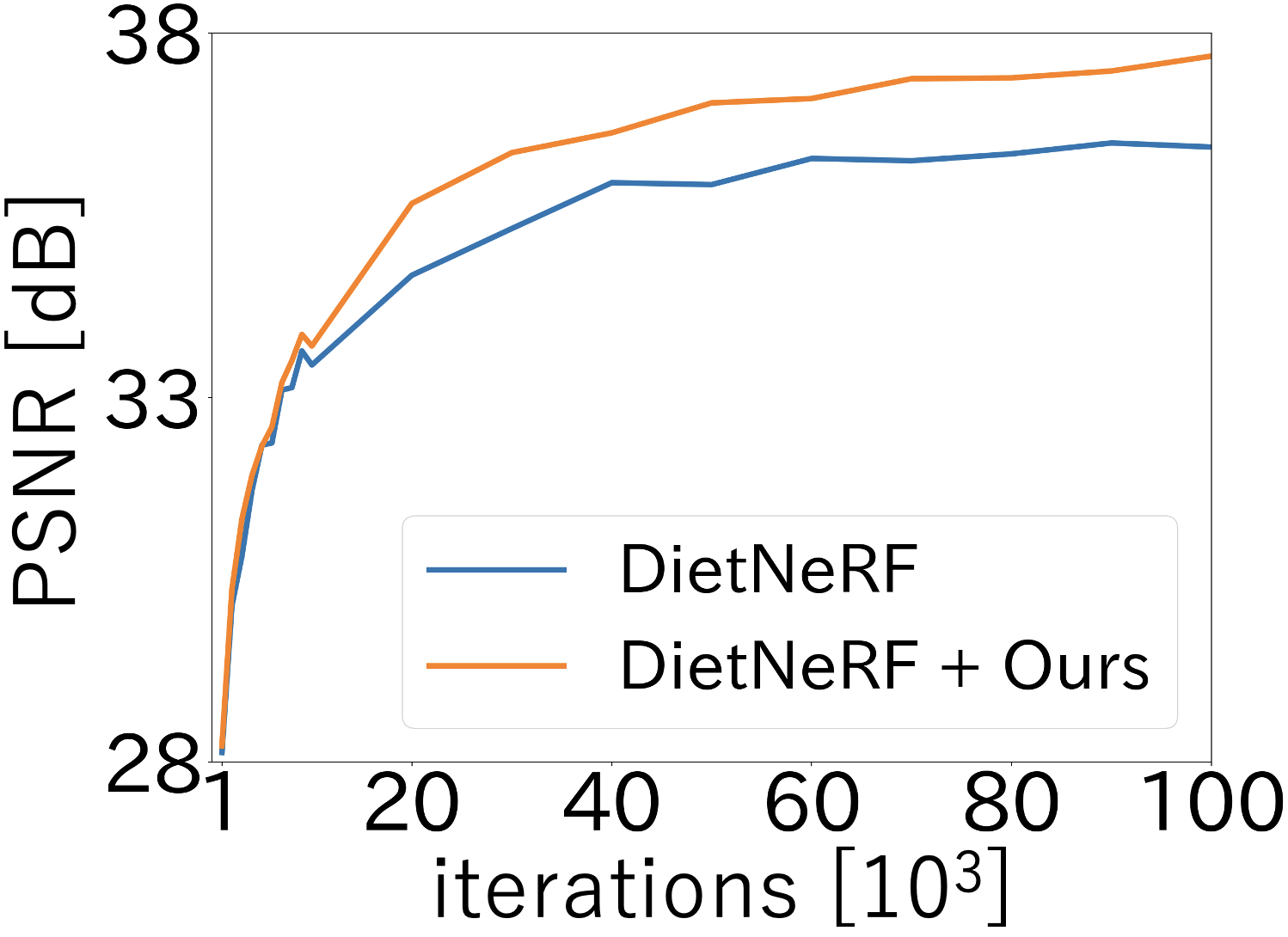}}
    \end{minipage}
    \hfill
    \begin{minipage}[b]{0.49\linewidth} 
      \centering
      \centerline{\includegraphics[width=\linewidth]{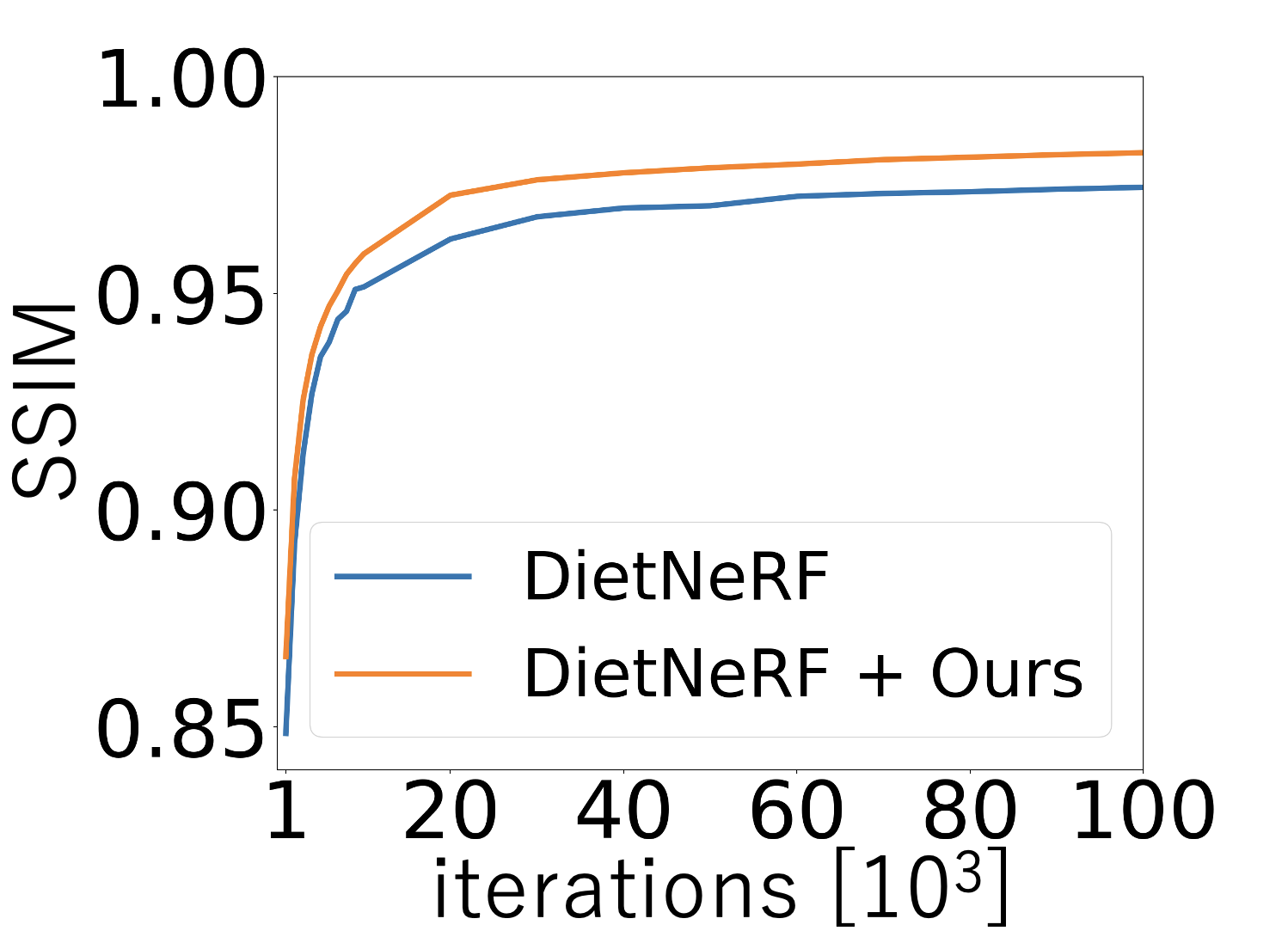}}
    \end{minipage}
    \centerline{Indoor scenes (from Replica Dataset~\cite{replica})}
\end{minipage}
\begin{minipage}[b]{0.49\linewidth}
    \begin{minipage}[b]{0.49\linewidth}
      \centering
      \centerline{\includegraphics[width=\linewidth]{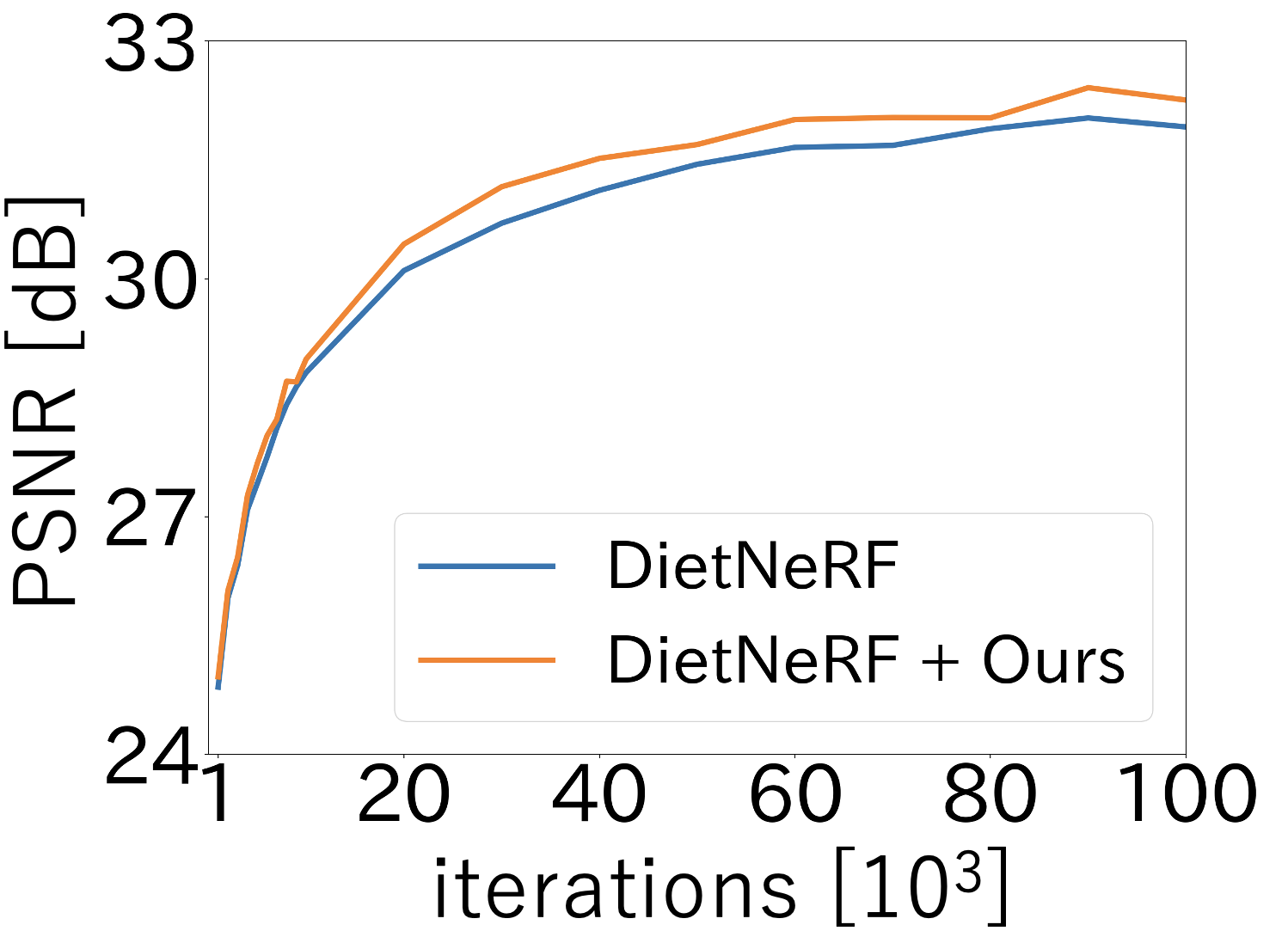}}
    \end{minipage}
    \hfill
    \begin{minipage}[b]{0.49\linewidth} 
      \centering
      \centerline{\includegraphics[width=\linewidth]{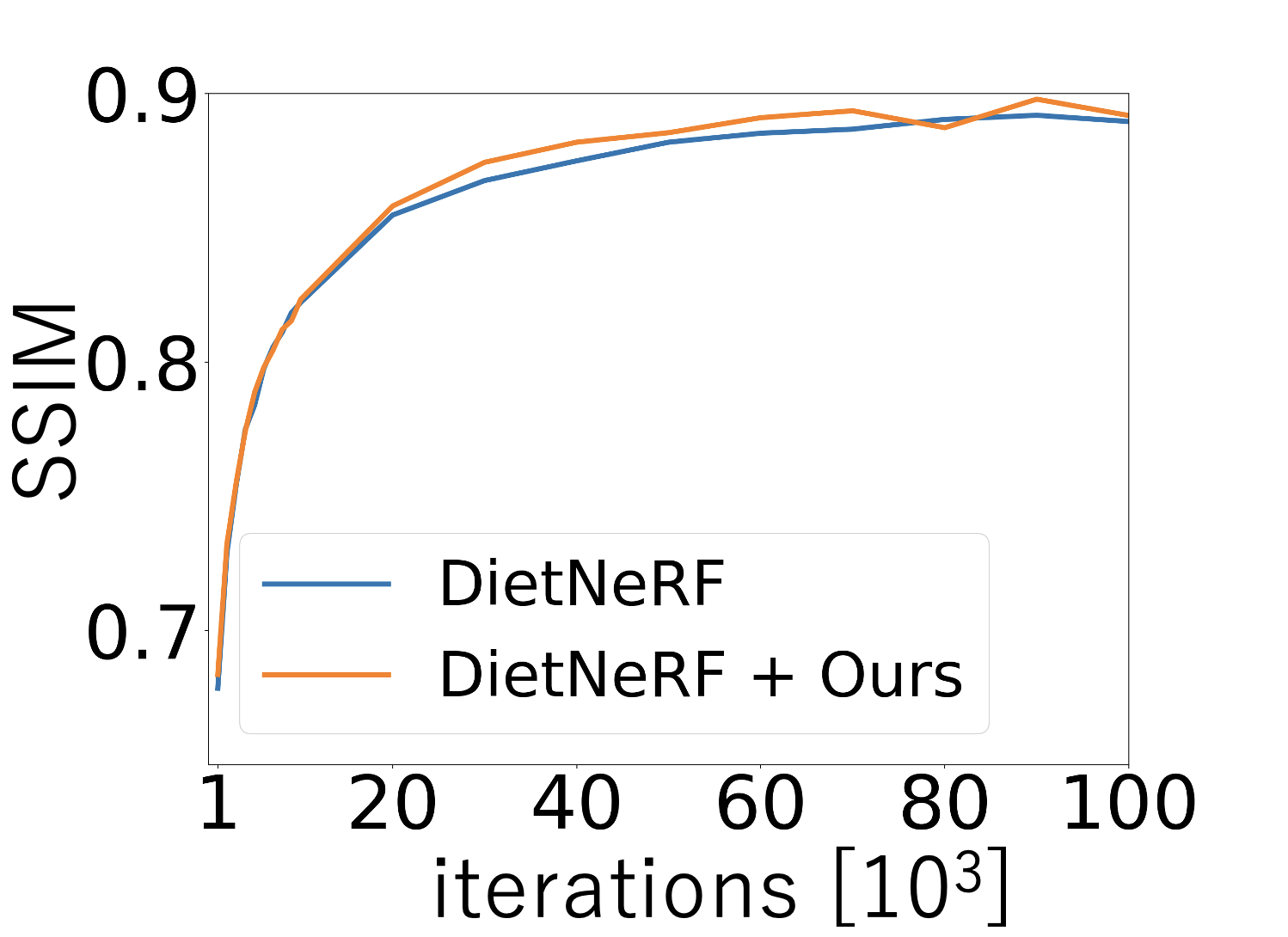}}
    \end{minipage}
    \centerline{Outdoor scenes (from SceneCity~\cite{scenecity})}
\end{minipage}
\centerline{(a) DietNeRF~\cite{dietnerf}}

\begin{minipage}[b]{0.49\linewidth}
    \begin{minipage}[b]{0.49\linewidth}
      \centering
      \centerline{\includegraphics[width=\linewidth]{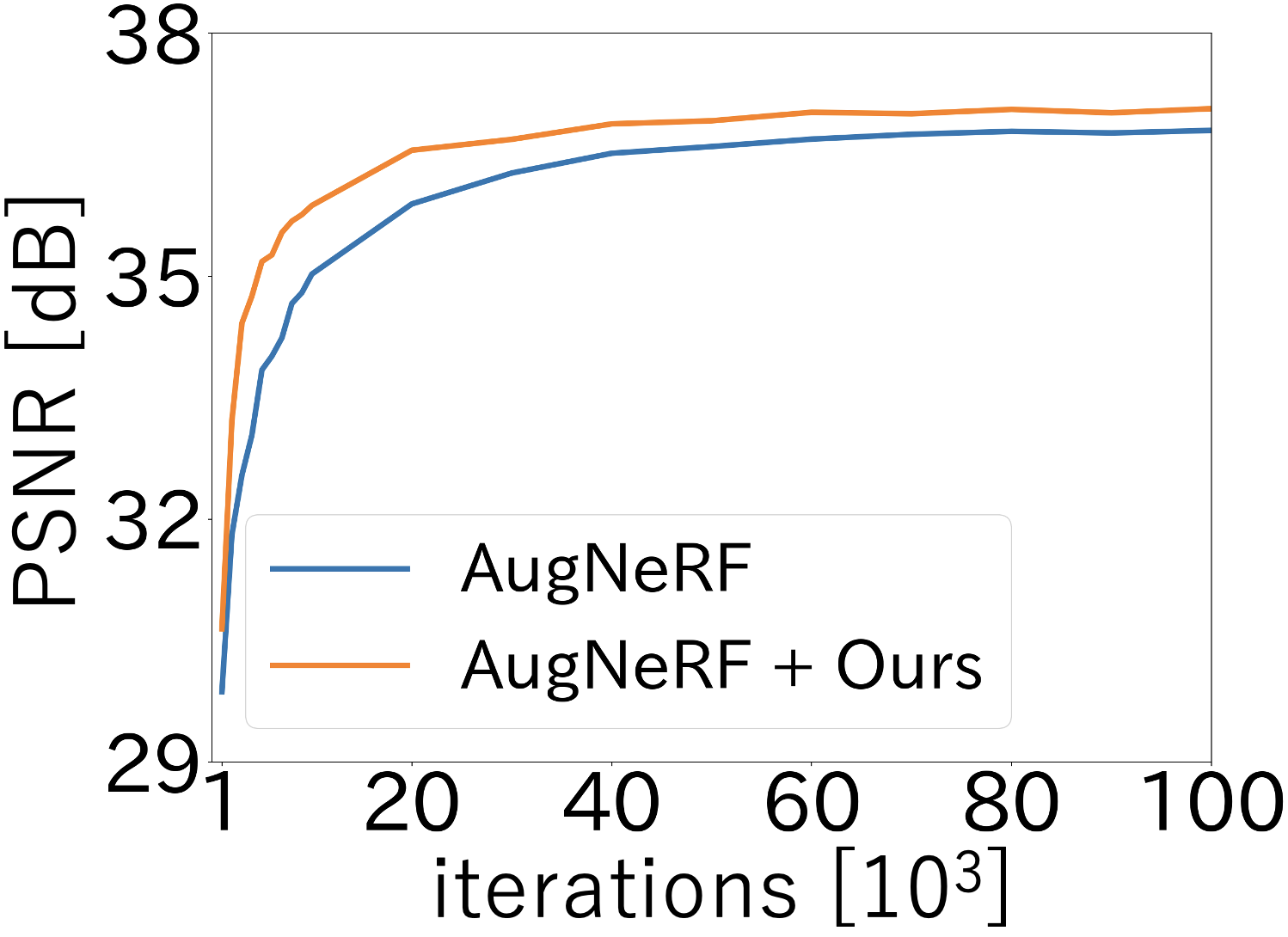}}
    \end{minipage}
    \hfill
    \begin{minipage}[b]{0.49\linewidth} 
      \centering
      \centerline{\includegraphics[width=\linewidth]{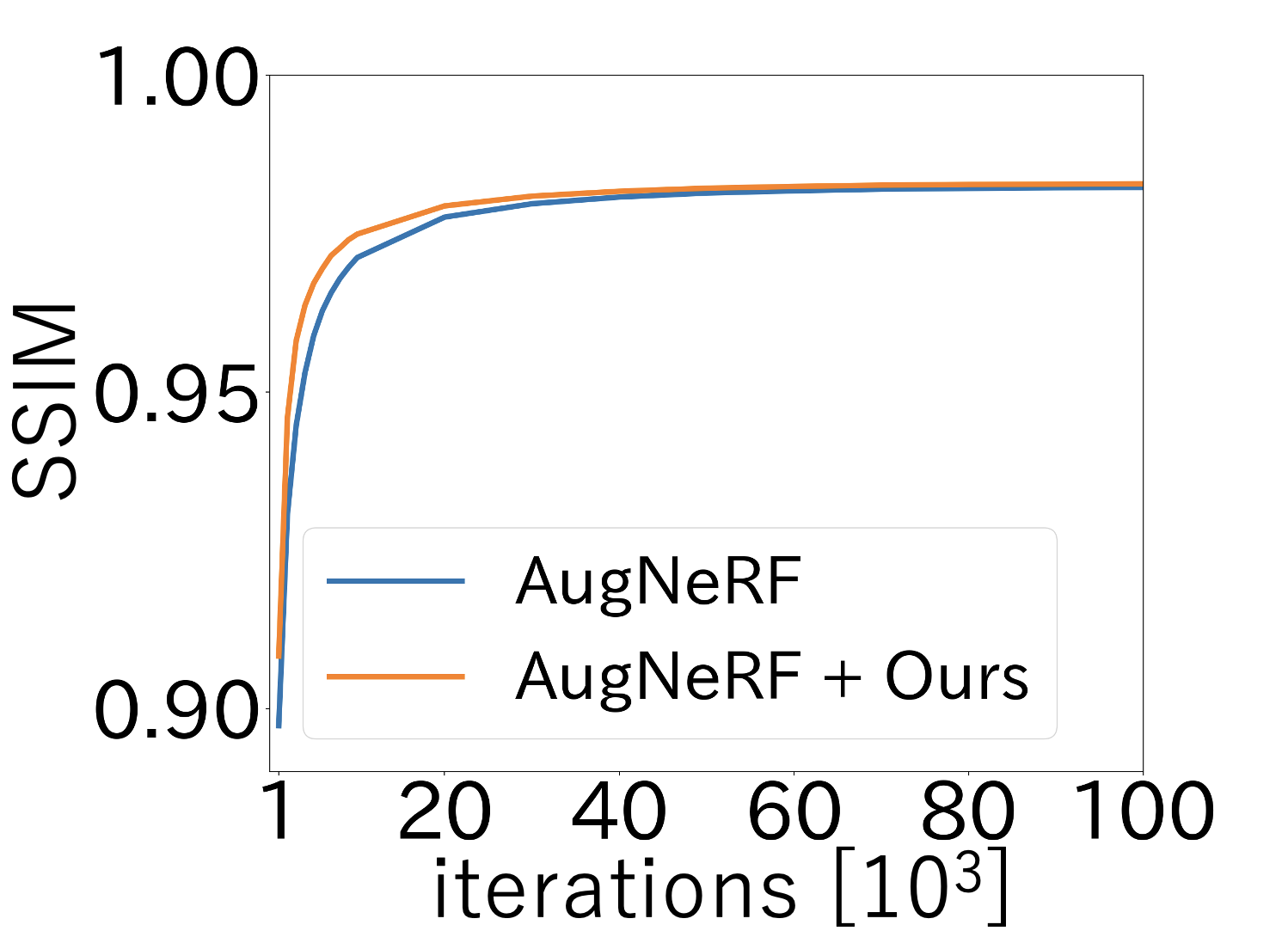}}
    \end{minipage}
    \centerline{Indoor scenes (from Replica Dataset~\cite{replica})}
\end{minipage}
\begin{minipage}[b]{0.49\linewidth}
    \begin{minipage}[b]{0.49\linewidth}
      \centering
      \centerline{\includegraphics[width=\linewidth]{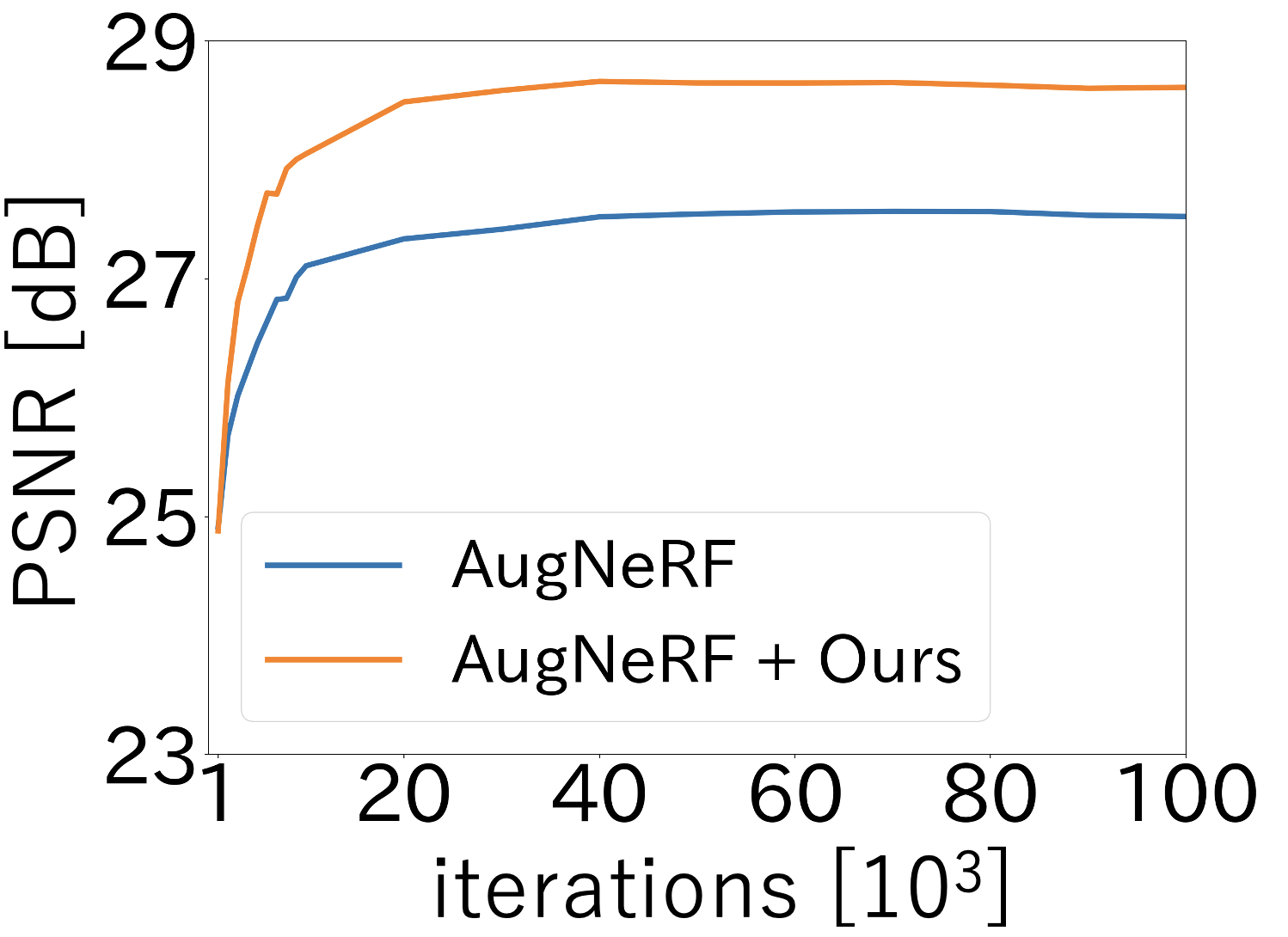}}
    \end{minipage}
    \hfill
    \begin{minipage}[b]{0.49\linewidth} 
      \centering
      \centerline{\includegraphics[width=\linewidth]{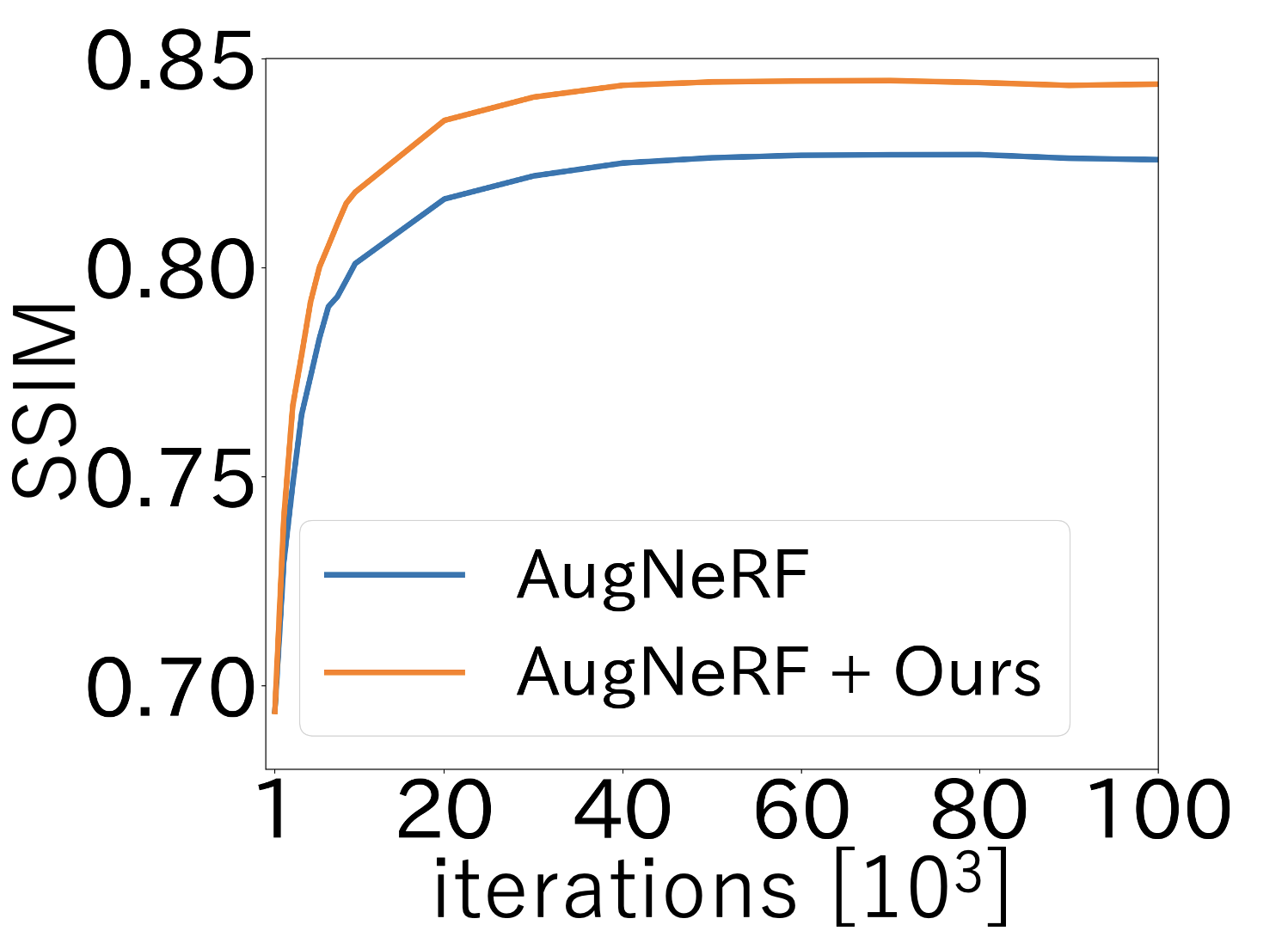}}
    \end{minipage}
    \centerline{Outdoor scenes (from SceneCity~\cite{scenecity})}
\end{minipage}
\centerline{(b) AugNeRF~\cite{chen2022augnerf}}

\begin{minipage}[b]{0.49\linewidth}
    \begin{minipage}[b]{0.49\linewidth}
      \centering
      \centerline{\includegraphics[width=\linewidth]{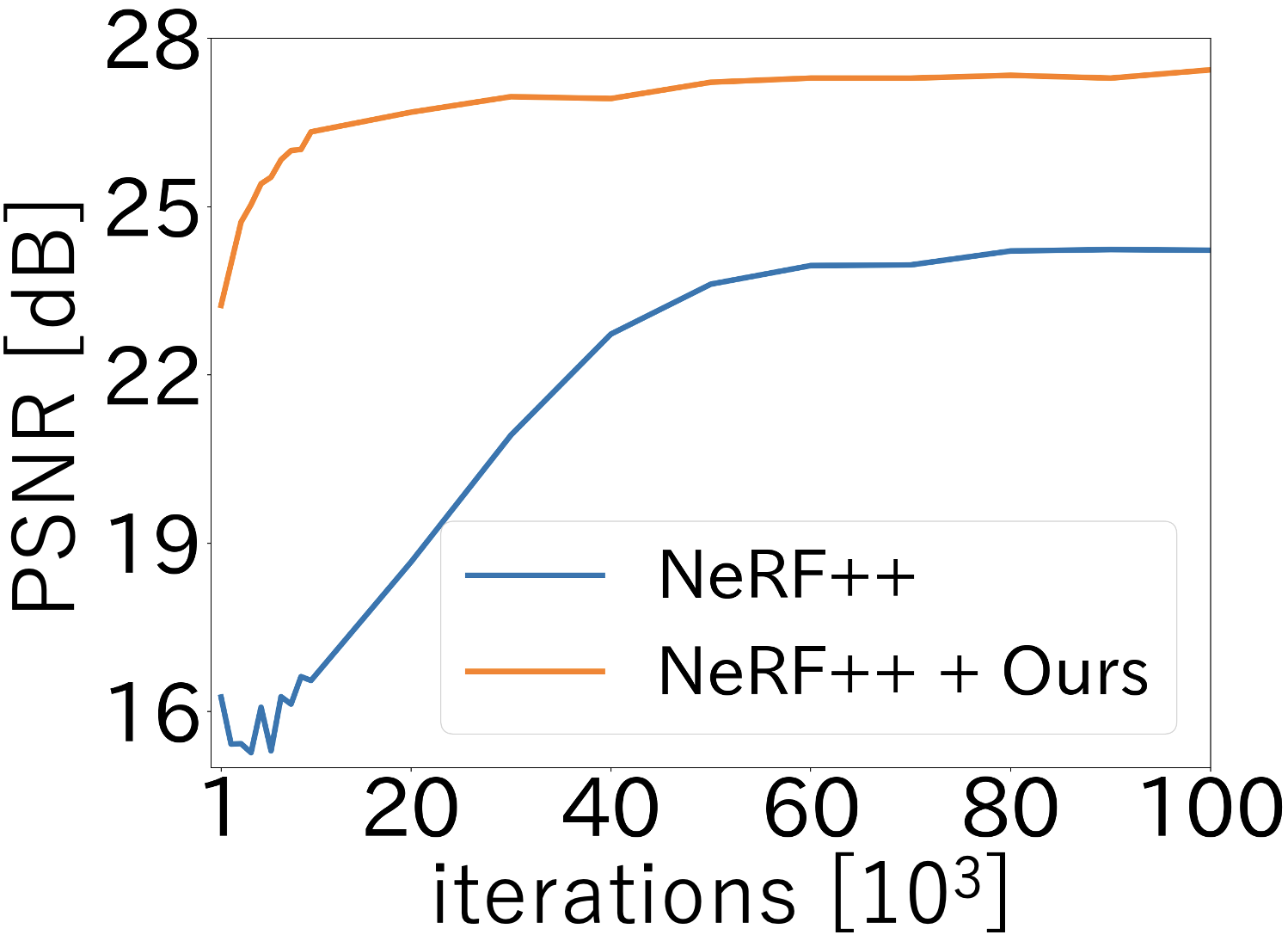}}
    \end{minipage}
    \hfill
    \begin{minipage}[b]{0.49\linewidth} 
      \centering
      \centerline{\includegraphics[width=\linewidth]{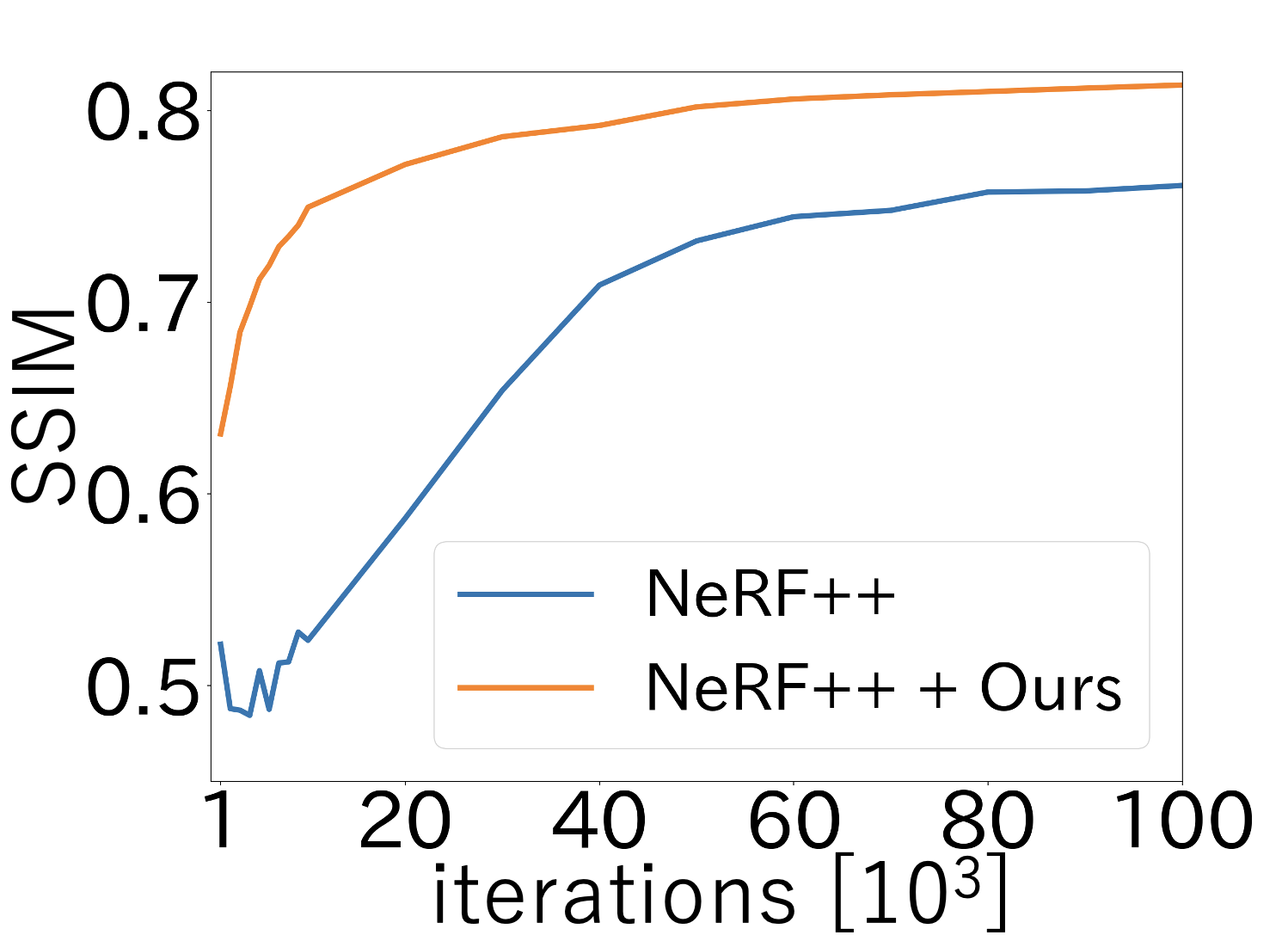}}
    \end{minipage}
    \centerline{scene\_1 (from SceneCity~\cite{scenecity})}
\end{minipage}
\begin{minipage}[b]{0.49\linewidth}
    \begin{minipage}[b]{0.49\linewidth}
      \centering
      \centerline{\includegraphics[width=\linewidth]{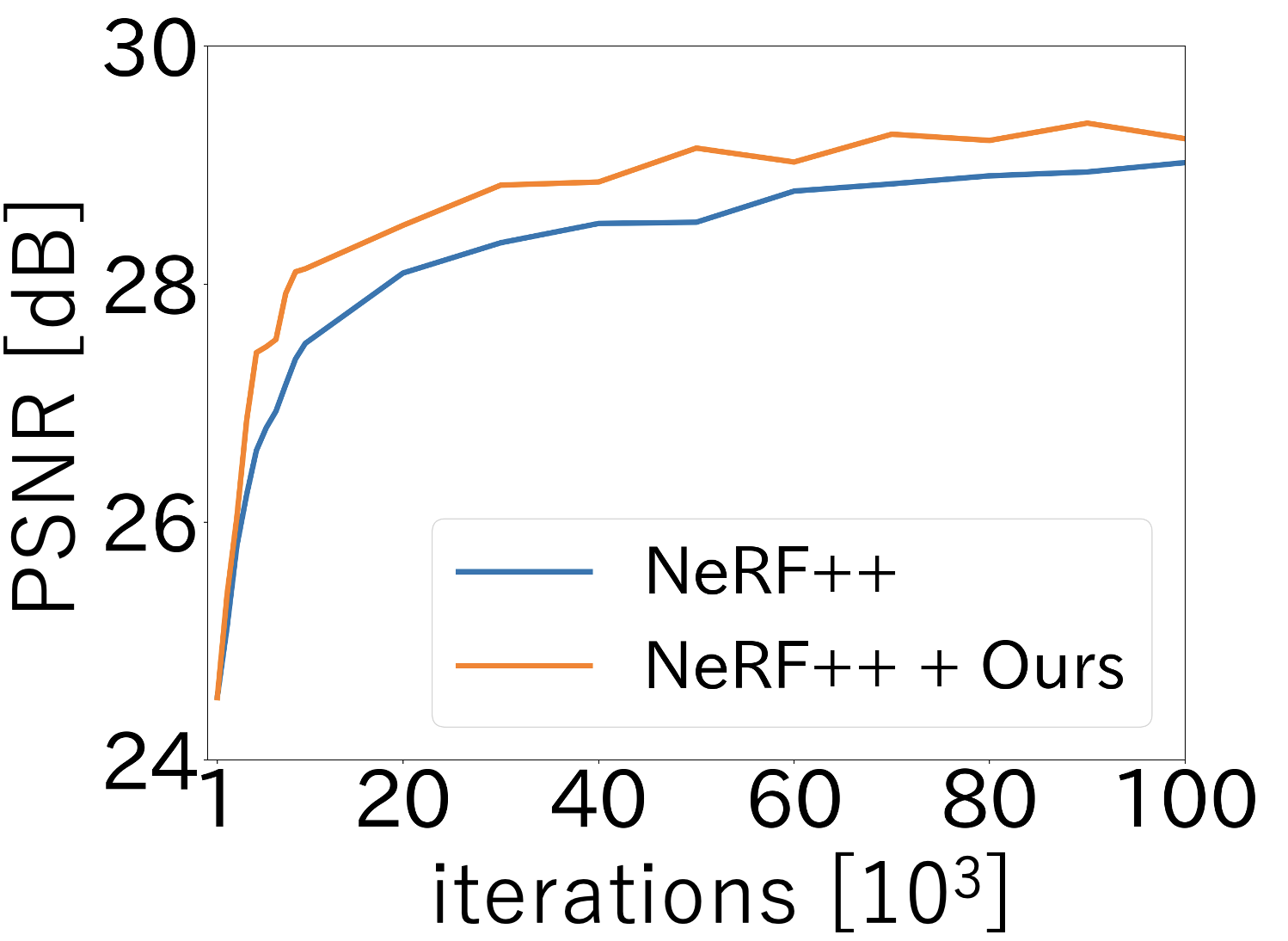}}
    \end{minipage}
    \hfill
    \begin{minipage}[b]{0.49\linewidth} 
      \centering
      \centerline{\includegraphics[width=\linewidth]{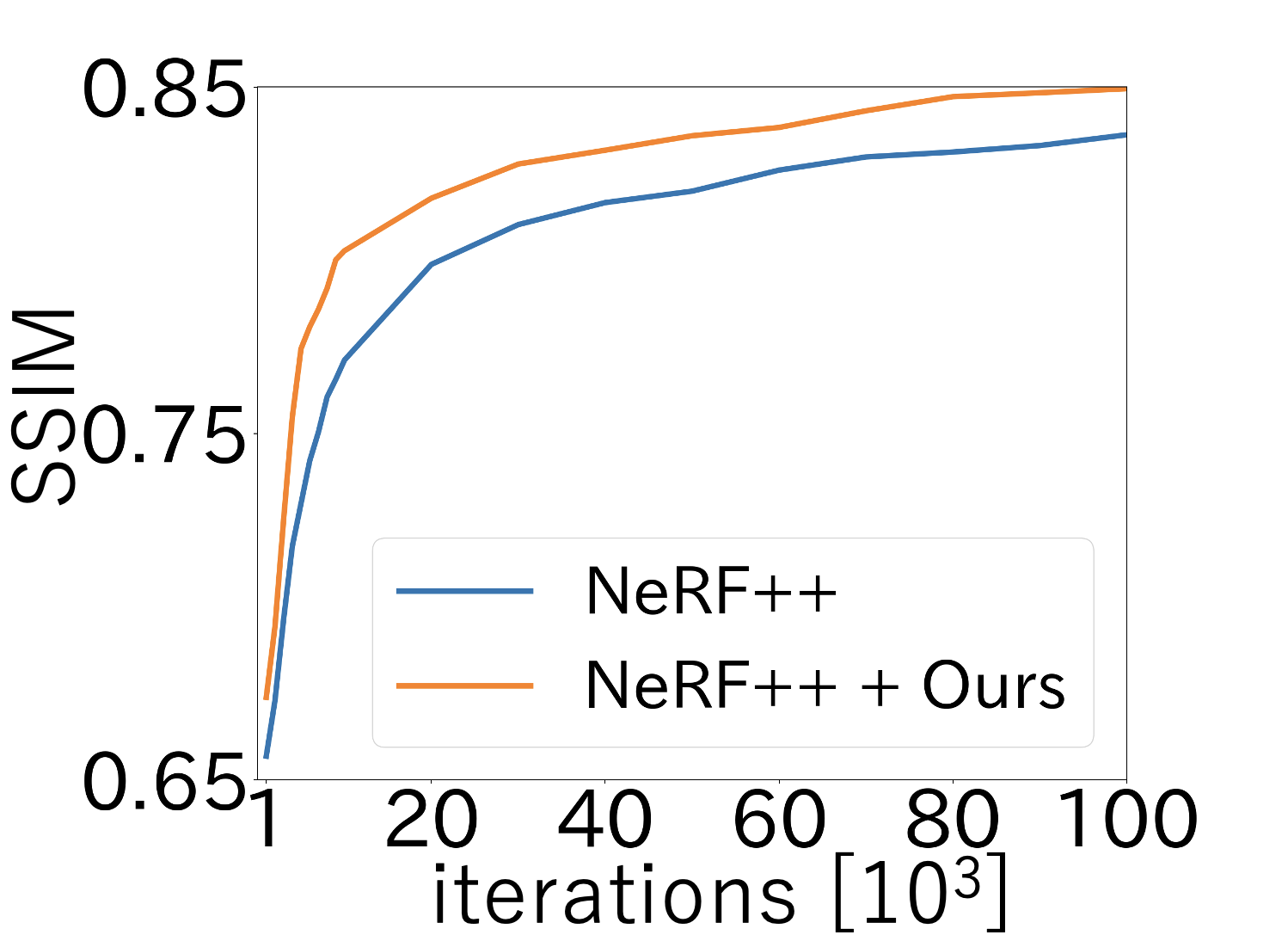}}
    \end{minipage}
    \centerline{scene\_2 (from SceneCity~\cite{scenecity})}
\end{minipage}
\centerline{(c) NeRF++~\cite{kaizhang2020}}
\caption{PSNR/SSIM curves of various advanced variants of NeRF combined with our proposed method.}
\label{fig:nerf_variants_psnr_graph}
\end{figure}

In the main paper, we showed the PSNR/SSIM scores at 100,000-th iteration. Here, we also show PSNR/SSIM curves averaged over indoor/outdoor datasets while training DietNeRF~\cite{dietnerf}, AugNeRF~\cite{chen2022augnerf}, and NeRF++~\cite{kaizhang2020} w/ or w/o our non-uniform sampling strategies. The results are shown in~\Fref{fig:nerf_variants_psnr_graph}. In summary, we observe the consistent improvement of learning curves with our method which shows both more efficient training and better accuracy at convergence.

\begin{figure*}[!t]
  \centering
  {\small
  \begin{minipage}[t!]{0.22\textwidth}
    \centering
    \includegraphics[height=14.5mm,width=29mm]{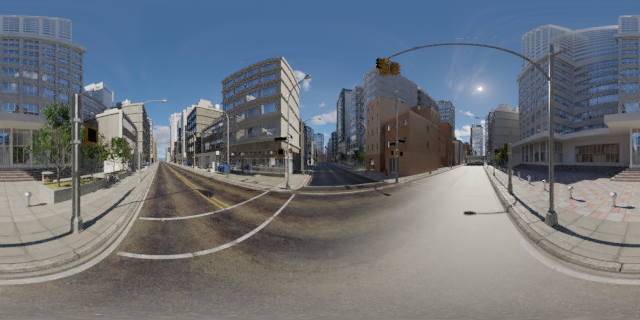}
    \vspace{-0.7cm}\center{Outdoor scene from Synth360}
  \end{minipage}
  \hfill
  \begin{minipage}[t!]{0.76\textwidth}
      \begin{minipage}[t]{0.32\textwidth}
        \centering
        \includegraphics[height=16mm,width=32mm]{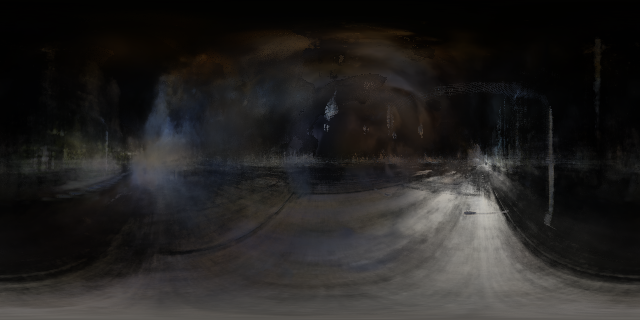}
        \vspace{-0.76cm}\center{\small foreground}
      \end{minipage}
      \begin{minipage}[t]{0.32\textwidth}
        \centering
        \includegraphics[height=16mm,width=32mm]{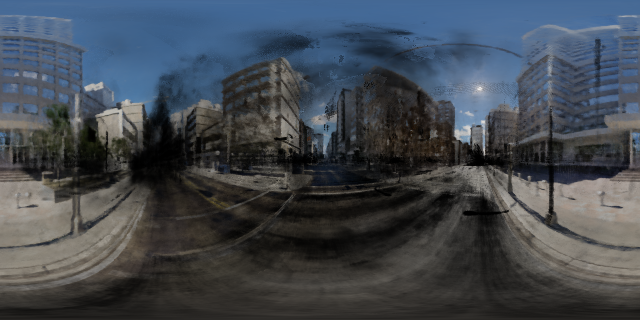}
        \vspace{-0.76cm}\center{\small background}
      \end{minipage}
      \begin{minipage}[t]{0.32\textwidth}
        \centering
        \includegraphics[height=16mm,width=32mm]{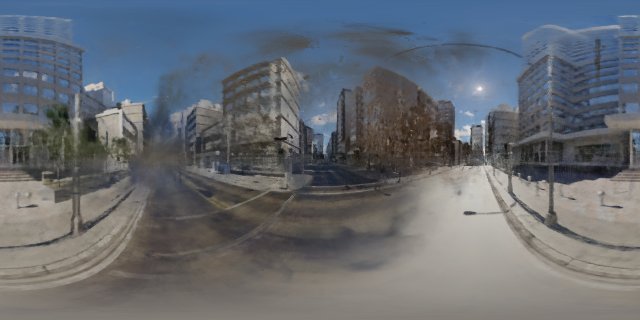}
        \vspace{-0.76cm}\center{\small NeRF++~\cite{kaizhang2020}}
      \end{minipage}
      \\
        \begin{minipage}[t]{0.32\textwidth}
        \centering
        \includegraphics[height=16mm,width=32mm]{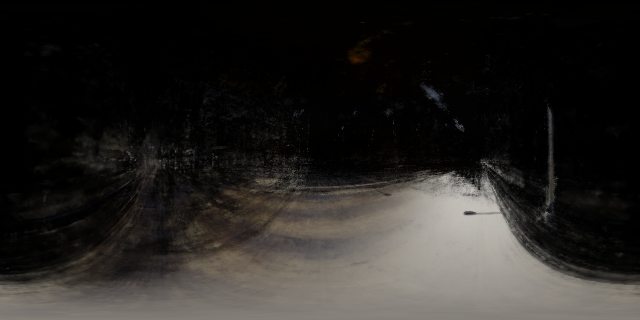}
        \vspace{-0.76cm}\center{\small foreground}
      \end{minipage}
        \begin{minipage}[t]{0.32\textwidth}
        \centering
        \includegraphics[height=16mm,width=32mm]{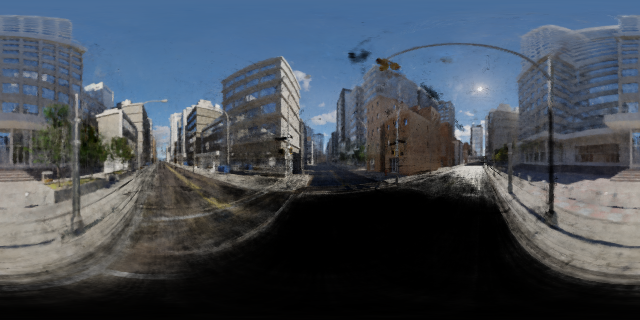}
        \vspace{-0.76cm}\center{\small background}
      \end{minipage}
        \begin{minipage}[t]{0.32\textwidth}
        \centering
        \includegraphics[height=16mm,width=32mm]{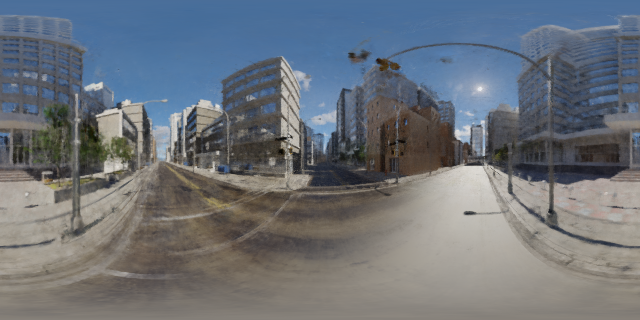}
        \vspace{-0.76cm}\center{\small NeRF++~\cite{kaizhang2020}+Ours}
      \end{minipage}
  \end{minipage}
  }
  \vspace{1mm}
\caption{Unsuccessful case of NeRF++~\cite{kaizhang2020}. We found that NeRF++ sometimes fails to separate between foreground and background. However, combining our proposed method often improved this failure case.}
\label{fig:experiment_3_nerfplusplus}
\end{figure*}

As mentioned, we found that NeRF++ often showed significant performance degradation when the foreground and background were not well separated (See~\Fref{fig:experiment_3_nerfplusplus}-top). Specifically, NeRF++ failed to be trained on indoor scenes in our Synth360 (\ie, Replica Dataset); therefore, we only show the result on outdoor scenes (\ie, SceneCity; scene\_1 and scene\_2). It is interesting to see that the n\"aive NeRF++ also failed to separate the foreground and background of the indoor dataset. However, combined with the proposed method, the learning of NeRF++ progressed appropriately on them. The characteristics of the individual algorithms are outside the scope of this work and will not be discussed further (\Fref{fig:experiment_3_nerfplusplus}-bottom).

\subsection{Qualitative Results}
\begin{figure*}[!t]
  \centering
  {\small
  \begin{minipage}[t]{0.27\textwidth}
    \centering
    \includegraphics[height=18mm,width=36mm]{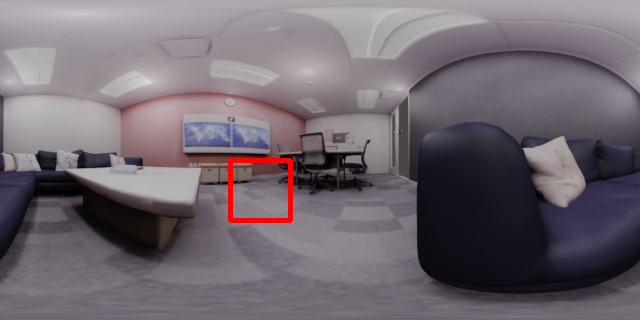}
    \vspace{-0.7cm}\center{Indoor dataset}
  \end{minipage}
  \hfill
  \begin{minipage}[t]{0.138\textwidth}
    \centering
    \includegraphics[height=18mm,width=18mm]{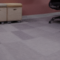}
    \vspace{-0.76cm}\center{\small Ground Truth}
  \end{minipage}
  \begin{minipage}[t]{0.138\textwidth}
    \centering
    \includegraphics[height=18mm,width=18mm]{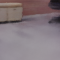}
    \vspace{-0.76cm}\center{\small DietNeRF~\cite{dietnerf}\\5,000 iterations}
  \end{minipage}
  \begin{minipage}[t]{0.138\textwidth}
    \centering
    \includegraphics[height=18mm,width=18mm]{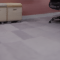}
    \vspace{-0.76cm}\center{\small DietNeRF~\cite{dietnerf}\\100,000 iterations}
  \end{minipage}
  \begin{minipage}[t]{0.138\textwidth}
    \centering
    \includegraphics[height=18mm,width=18mm]{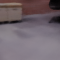}
    \vspace{-0.76cm}\center{\small Ours\\5,000 iterations}
  \end{minipage}
  \begin{minipage}[t]{0.138\textwidth}
    \centering
    \includegraphics[height=18mm,width=18mm]{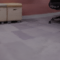}
    \vspace{-0.76cm}\center{\small Ours\\100,000 iterations}
  \end{minipage}
  \\
  \begin{minipage}[t]{0.27\textwidth}
    \centering
    \includegraphics[height=17.8mm,width=35.6mm]{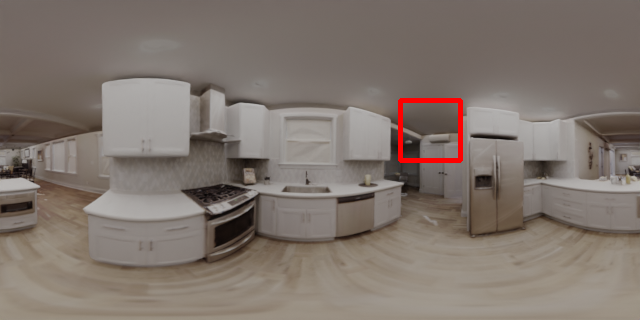}
    \vspace{-0.7cm}\center{Indoor dataset}
  \end{minipage}
  \hfill
  \begin{minipage}[t]{0.138\textwidth}
    \centering
    \includegraphics[height=18mm,width=18mm]{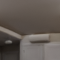}
    \vspace{-0.76cm}\center{\small Ground Truth}
  \end{minipage}
  \begin{minipage}[t]{0.138\textwidth}
    \centering
    \includegraphics[height=18mm,width=18mm]{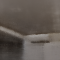}
    \vspace{-0.76cm}\center{\small AugNeRF~\cite{chen2022augnerf}\\5,000 iterations}
  \end{minipage}
  \begin{minipage}[t]{0.138\textwidth}
    \centering
    \includegraphics[height=18mm,width=18mm]{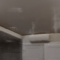}
    \vspace{-0.76cm}\center{\small AugNeRF~\cite{chen2022augnerf}\\100,000 iterations}
  \end{minipage}
  \begin{minipage}[t]{0.138\textwidth}
    \centering
    \includegraphics[height=18mm,width=18mm]{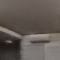}
    \vspace{-0.76cm}\center{\small Ours\\5,000 iterations}
  \end{minipage}
  \begin{minipage}[t]{0.138\textwidth}
    \centering
    \includegraphics[height=18mm,width=18mm]{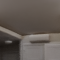}
    \vspace{-0.76cm}\center{\small Ours\\100,000 iterations}
  \end{minipage}\\

  \begin{minipage}[t]{0.27\textwidth}
    \centering
    \includegraphics[height=17.8mm,width=35.6mm]{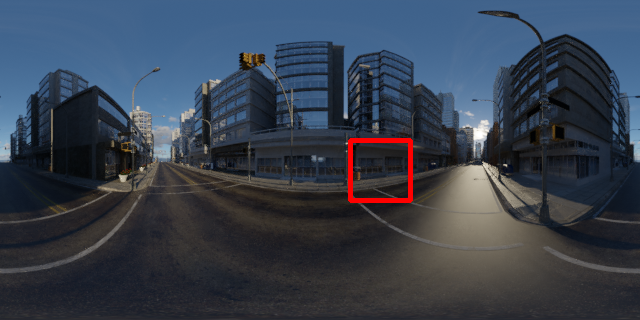}
    \vspace{-0.7cm}\center{Outdoor dataset}
  \end{minipage}
  \hfill
  \begin{minipage}[t]{0.138\textwidth}
    \centering
    \includegraphics[height=18mm,width=18mm]{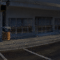}
    \vspace{-0.76cm}\center{\small Ground Truth}
  \end{minipage}
  \begin{minipage}[t]{0.138\textwidth}
    \centering
    \includegraphics[height=18mm,width=18mm]{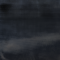}
    \vspace{-0.76cm}\center{\small NeRF++~\cite{kaizhang2020}\\5,000 iterations}
  \end{minipage}
  \begin{minipage}[t]{0.138\textwidth}
    \centering
    \includegraphics[height=18mm,width=18mm]{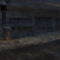}
    \vspace{-0.76cm}\center{\small NeRF++~\cite{kaizhang2020}\\100,000 iterations}
  \end{minipage}
  \begin{minipage}[t]{0.138\textwidth}
    \centering
    \includegraphics[height=18mm,width=18mm]{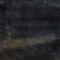}
    \vspace{-0.76cm}\center{\small Ours\\5,000 iterations}
  \end{minipage}
  \begin{minipage}[t]{0.138\textwidth}
    \centering
    \includegraphics[height=18mm,width=18mm]{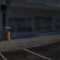}
    \vspace{-0.76cm}\center{\small Ours\\100,000 iterations}
  \end{minipage}\\
  }
\caption{Qualitative comparison using DietNeRF~\cite{dietnerf}, AugNeRF~\cite{chen2022augnerf}, and NeRF++~\cite{kaizhang2020} on Synth360.}
\label{fig:nerf_variant_image}
\end{figure*}

The qualitative results of DietNeRF~\cite{dietnerf}, AugNeRF~\cite{chen2022augnerf}, and NeRF++~\cite{kaizhang2020} w/ and w/o our non-uniform sampling strategies are shown in \Fref{fig:nerf_variant_image}. We also observe our proposed method is also effective for advanced variants of NeRF: better high-frequency texture recovery and less artifacts.

\section{Detailed Behavior of Non-Uniform Sampling Strategies}
So far, we have evaluated our method based on quantitative and qualitative comparisons on an entire image and validated that our non-uniform sampling strategies are effective for improving both learning curves and the reconstruction quality at the convergence despite its simplicity. However, since the motivation of the proposed method is to dynamically change the sampling probability according to the amount of sphere-to-plane projection distortion at high latitudes and the content of 360\textdegree~images with a wide field of view, this section verifies whether our strategies actually work as intended.

\begin{figure}[!t]
\begin{minipage}[b]{0.49\linewidth}
  \centering
  \centerline{\includegraphics[width=\linewidth]{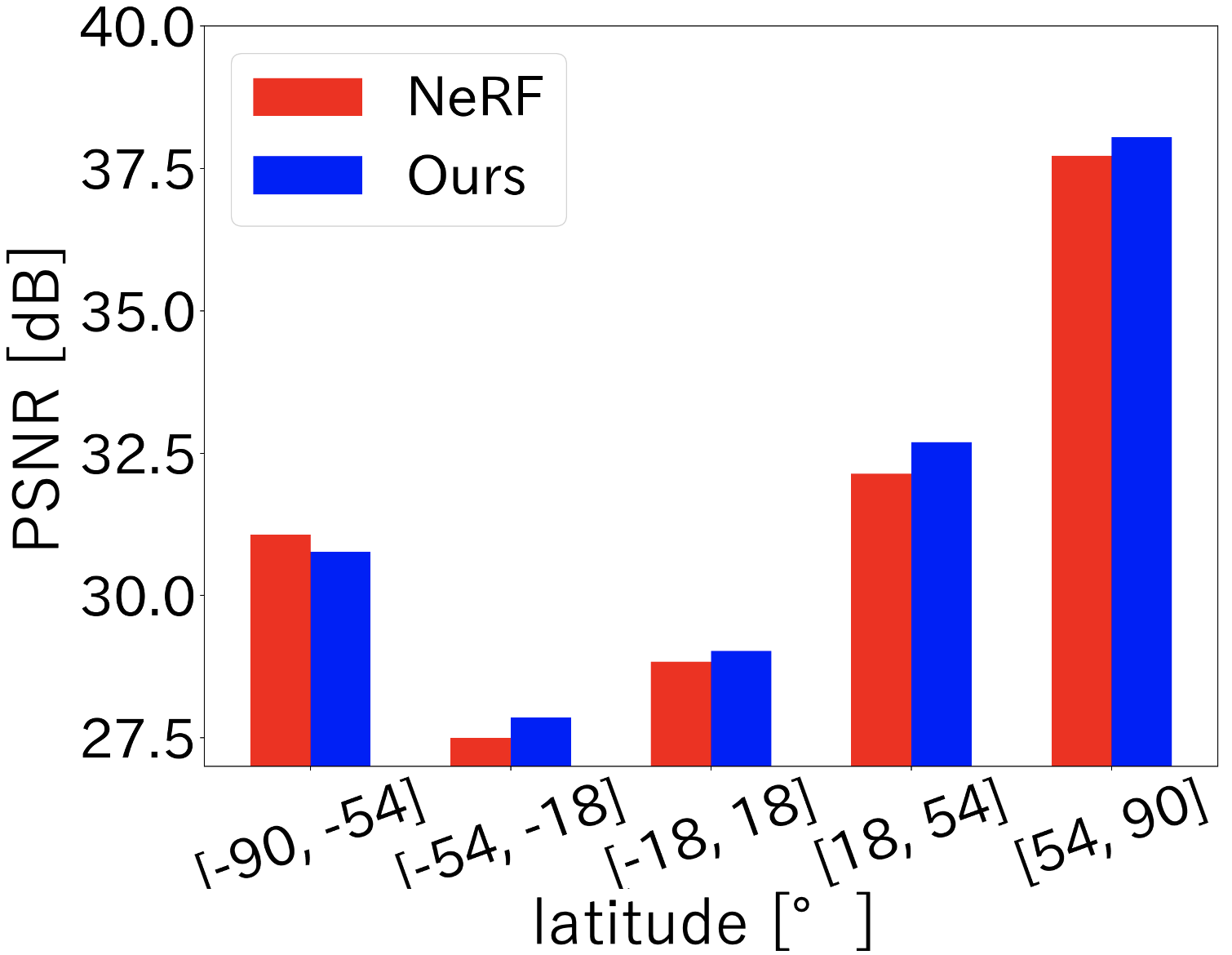}}
  \centerline{Indoor scenes (from Replica Dataset~\cite{replica})}
\end{minipage}
\hfill
\begin{minipage}[b]{0.49\linewidth} 
  \centering
  \centerline{\includegraphics[width=\linewidth]{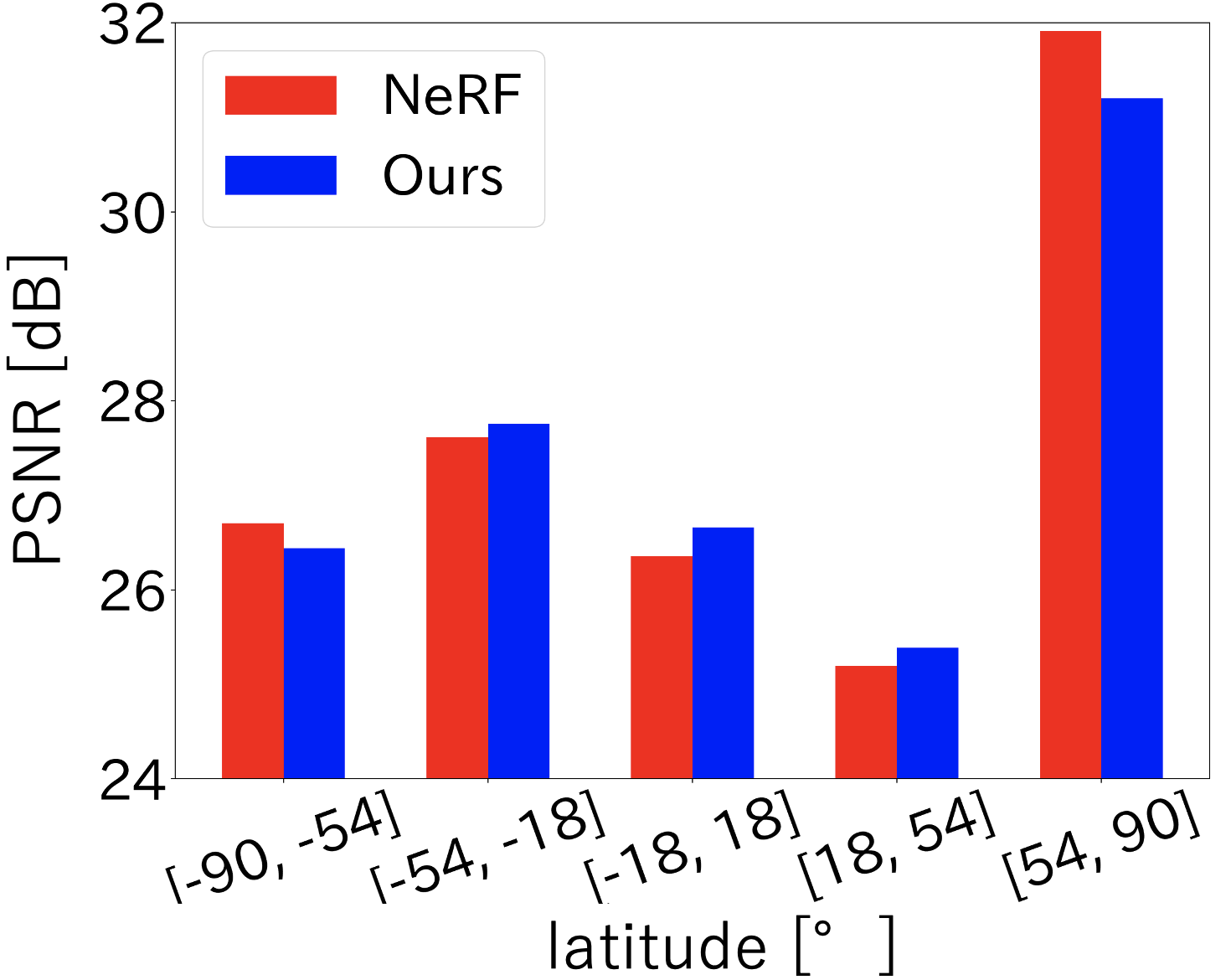}}
  \centerline{Outdoor scenes (from SceneCity~\cite{scenecity})}
\end{minipage}
\caption{Comparison by PSNR for each region of latitude. Latitude is assigned as -90\textdegree~at the bottom of the image and 90\textdegree~at the top of the image. Thus, [-90, -54] and [54, 90] are high latitude regions and [-18, 18] are low latitude regions.}
\label{fig:experiment4_distortion}
\end{figure}

First, to validate the behavior of our distortion-aware ray sampling, we divided the ERP coordinate into five regions by latitude and made quantitative comparisons in each region as shown in \Fref{fig:experiment4_distortion}. As expected, since our distortion-aware sampling strategy assigns a higher sampling probability in lower-latitude regions, the reconstruction accuracy in low-latitude regions improved while one in high-latitude regions often slightly degraded. It is important to note that PSNR is computed in ERP coordinates; therefore, the reconstruction accuracy at high latitudes, where the amount of information is far less than at low latitudes, is less important for most practical applications.

Second, to validate the behavior of our content-aware ray sampling, we cropped high-frequency and low-frequency texture crops of 60$\times$60 and evaluated our method on each crop individually. We observed that our content-aware ray sampling probability assigns large probability values to regions with many edges. Therefore, we considered regions with many edges to be high-frequency regions and regions with few edges to be low-frequency regions. Specifically, a 3$\times$3 laplacian filter was applied to the ERP image, and crops with large outputs were defined as high-frequency texture regions, while regions with small values were defined as low-frequency texture regions. One low-frequency texture region and one high-frequency texture region were extracted for each test image. As shown in \Tref{table:experiment4_content}, our proposed method fairly improved the reconstruction accuracy for the high-frequency texture crops because our content-aware ray sampling strategy increased the number of samples in challenging high-frequency texture regions. On the other hand, though the proposed method reduced the number of samples for the low-frequency crops, which are easy to learn, its accuracy degradation was found to be small.

\begin{table}[!t]
    \caption{Quantitative comparison (PSNR) using NeRF~\cite{nerf} and Ours on low-frequency texture and high-frequency texture regions. The best is highlighted.}
    \vspace{5pt}
    \label{table:experiment4_content}
    \centering
    \begin{tabular}{c|r|r|r|r}
    & \multicolumn{2}{|c|}{Indoor scenes (Synth360)} &  \multicolumn{2}{c}{Outdoor scenes (Synth360)}\\
    Method & low-frequency & high-frequency & low-frequency & high-frequency \\\hline
    NeRF~\cite{nerf} & 44.63 & 29.68 & \textbf{38.28} & 23.51 \\
    Ours & \textbf{44.85} & \textbf{31.63} & 37.42 & \textbf{24.86} \\
    \end{tabular}
\end{table}

\begin{figure*}[!t]
  \centering
  {\small

  \begin{minipage}[t]{0.27\textwidth}
    \centering
    \includegraphics[height=18mm,width=36mm]{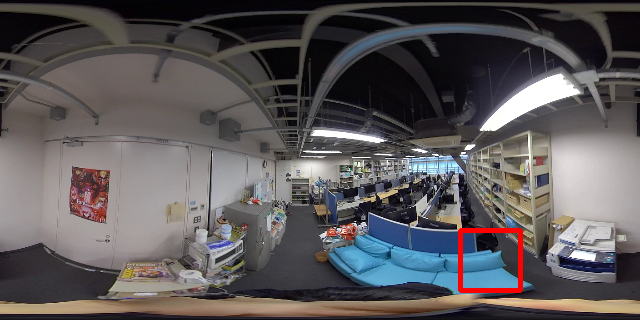}
    \vspace{-0.7cm}\center{indoor scene}
  \end{minipage}
  \hfill
  \begin{minipage}[t]{0.138\textwidth}
    \centering
    \includegraphics[height=18mm,width=18mm]{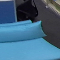}
    \vspace{-0.76cm}\center{\small Ground Truth}
  \end{minipage}
  \begin{minipage}[t]{0.138\textwidth}
    \centering
    \includegraphics[height=18mm,width=18mm]{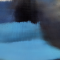}
    \vspace{-0.76cm}\center{\small DietNeRF~\cite{dietnerf}\\5,000 iterations}
  \end{minipage}
  \begin{minipage}[t]{0.138\textwidth}
    \centering
    \includegraphics[height=18mm,width=18mm]{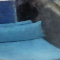}
    \vspace{-0.76cm}\center{\small DietNeRF~\cite{dietnerf}\\100,000 iterations}
  \end{minipage}
  \begin{minipage}[t]{0.138\textwidth}
    \centering
    \includegraphics[height=18mm,width=18mm]{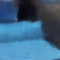}
    \vspace{-0.76cm}\center{\small Ours\\5,000 iterations}
  \end{minipage}
  \begin{minipage}[t]{0.138\textwidth}
    \centering
    \includegraphics[height=18mm,width=18mm]{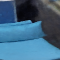}
    \vspace{-0.76cm}\center{\small Ours\\100,000 iterations}
  \end{minipage}

  \begin{minipage}[t]{0.27\textwidth}
    \centering
    \includegraphics[height=18mm,width=36mm]{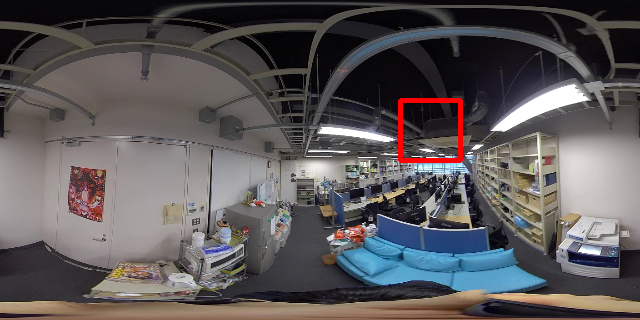}
    \vspace{-0.7cm}\center{indoor scene}
  \end{minipage}
  \hfill
  \begin{minipage}[t]{0.138\textwidth}
    \centering
    \includegraphics[height=18mm,width=18mm]{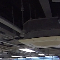}
    \vspace{-0.76cm}\center{\small Ground Truth}
  \end{minipage}
  \begin{minipage}[t]{0.138\textwidth}
    \centering
    \includegraphics[height=18mm,width=18mm]{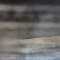}
    \vspace{-0.76cm}\center{\small AugNeRF~\cite{chen2022augnerf}\\5,000 iterations}
  \end{minipage}
  \begin{minipage}[t]{0.138\textwidth}
    \centering
    \includegraphics[height=18mm,width=18mm]{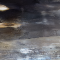}
    \vspace{-0.76cm}\center{\small AugNeRF~\cite{chen2022augnerf}\\100,000 iterations}
  \end{minipage}
  \begin{minipage}[t]{0.138\textwidth}
    \centering
    \includegraphics[height=18mm,width=18mm]{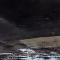}
    \vspace{-0.76cm}\center{\small Ours\\5,000 iterations}
  \end{minipage}
  \begin{minipage}[t]{0.138\textwidth}
    \centering
    \includegraphics[height=18mm,width=18mm]{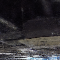}
    \vspace{-0.76cm}\center{\small Ours\\100,000 iterations}
  \end{minipage}
  \\
  

  \begin{minipage}[t]{0.27\textwidth}
    \centering
    \includegraphics[height=17.8mm,width=35.6mm]{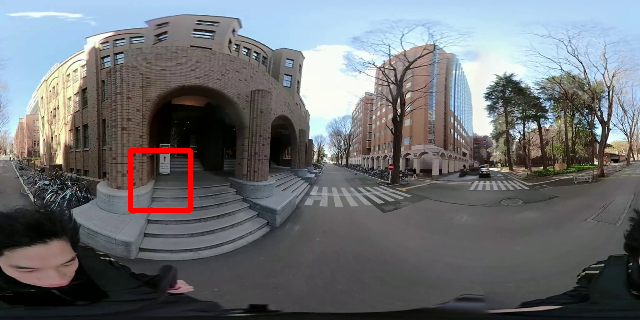}
    \vspace{-0.7cm}\center{outdoor scene}
  \end{minipage}
  \hfill
  \begin{minipage}[t]{0.138\textwidth}
    \centering
    \includegraphics[height=18mm,width=18mm]{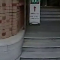}
    \vspace{-0.76cm}\center{\small Ground Truth}
  \end{minipage}
  \begin{minipage}[t]{0.138\textwidth}
    \centering
    \includegraphics[height=18mm,width=18mm]{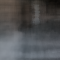}
    \vspace{-0.76cm}\center{\small NeRF++~\cite{kaizhang2020}\\5,000 iterations}
  \end{minipage}
  \begin{minipage}[t]{0.138\textwidth}
    \centering
    \includegraphics[height=18mm,width=18mm]{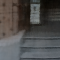}
    \vspace{-0.76cm}\center{\small NeRF++~\cite{kaizhang2020}\\100,000 iterations}
  \end{minipage}
  \begin{minipage}[t]{0.138\textwidth}
    \centering
    \includegraphics[height=18mm,width=18mm]{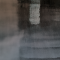}
    \vspace{-0.76cm}\center{\small Ours\\5,000 iterations}
  \end{minipage}
  \begin{minipage}[t]{0.138\textwidth}
    \centering
    \includegraphics[height=18mm,width=18mm]{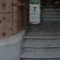}
    \vspace{-0.76cm}\center{\small Ours\\100,000 iterations}
  \end{minipage}\\

  }
\caption{Qualitative comparison on our real-world indoor and outdoor scenes.}
\label{fig:realworld_image}
\end{figure*}

\section{Qualitative comparison using advanced variants of NeRF on real-world scenes}
The qualitative results of DietNeRF~\cite{dietnerf}, AugNeRF~\cite{chen2022augnerf}, and NeRF++~\cite{kaizhang2020} combined with our proposed method on real-world indoor and outdoor scenes are shown in \Fref{fig:realworld_image}. We found that our proposed method works well for real-world scenes and improves the synthesis quality in the high-frequency texture regions.


\newpage
\bibliography{egbib}